%% file: arxiv.tex
\documentclass{article}
\PassOptionsToPackage{table,xcdraw}{xcolor}
\usepackage{arxiv}
\usepackage{times}

\usepackage[utf8]{inputenc}
\usepackage[T1]{fontenc}
\usepackage{natbib}
\usepackage{hyperref}
\usepackage{url}
\usepackage{booktabs}
\usepackage{amsfonts}
\usepackage{nicefrac}
\usepackage{microtype}
\usepackage{enumitem}
\usepackage{graphicx}
\usepackage{multirow}
\usepackage{xcolor}
\usepackage{amsmath,amssymb,amsthm,mathtools}
\usepackage{bm}
\usepackage{tabularx}
\usepackage{placeins}

\usepackage[font=small,labelfont=bf,skip=4pt]{caption}

\usepackage{longtable}

\usepackage{float}
\floatstyle{plaintop}
\restylefloat{table}

\newtheorem{proposition}{Proposition}

\theoremstyle{definition}
\newtheorem{definition}{Definition}
\newtheorem{remark}{Remark}
\newtheorem{assumption}{Assumption}

\newcommand{\E}{\mathbb{E}}
\newcommand{\R}{\mathbb{R}}

\newcommand{\Dfit}{\mathcal{D}_{\mathrm{fit}}}
\newcommand{\Dcal}{\mathcal{D}_{\mathrm{calib}}}
\newcommand{\Dtest}{\mathcal{D}_{\mathrm{test}}}

\newcommand{\cT}{\mathcal{D}_{\mathrm{test}}}

\title{Risk-Controlled DFT Selection for Machine-Learned Interatomic Potentials}
\date{}
\renewcommand{\shorttitle}{Risk-Controlled DFT Selection for MLIPs}

\author{
\textbf{Kun Ryu}$^{1,\dagger}$
\qquad
\textbf{Yeo Jin Jung}$^{2,\dagger}$
\qquad
\textbf{Shyue Ping Ong}$^{3}$
\qquad
\textbf{Claire Donnat}$^{2}$
\qquad
\textbf{Ying Shirley Meng}$^{1,*}$
\\[0.5ex]
$^{1}$Pritzker School of Molecular Engineering, The University of Chicago
\\
$^{2}$Department of Statistics, The University of Chicago
\\
$^{2}$Department of Material Science and Engineering, National University of Singapore
\\[0.5ex]
$^{\dagger}$Equal contribution
\qquad
$^{*}$Corresponding author
}

\hypersetup{
pdftitle={Risk-Controlled DFT Selection for Machine-Learned Interatomic Potentials},
pdfauthor={Kun Ryu, Yeo Jin Jung, Ying Shirley Meng, Claire Donnat},
}

\input{tables/values}

\input{tables/values_restored}

\begin{document}
\raggedbottom
\maketitle

\begin{abstract}
\input{texts/abstract}

\end{abstract}

\section{Introduction}
\input{texts/intro.tex}

\section{Method}
\input{texts/setup}
\input{texts/error_predictor}

\input{texts/crc}

\section{Experiments}\label{sec:experiments}
\input{texts/experiments}

\section{Conclusion}
\input{texts/conclusion}

\section*{AI Use Statement}
In this work, we used generative AI tools to assist in the writing of proofs, provide feedback on research methodology and experiments, implement methods, and
interpret results. We have not used generative AI tools to generate synthetic data, develop the theoretical framework, formulate the mathematical claims,
propose hypotheses, design the experimental methodology, translate, or perform
qualitative data analysis. Additionally, we used generative AI tools to edit software code, draft parts of this paper, produce and refine figures, edit the
manuscript for readability, and search for related literature. The research problem, the proposed approach, and all claims are the authors' own: every proof
was verified independently of the assistant that proposed it, all AI-assisted code was read and its outputs reproduced, and all references were checked against
their sources. We take full responsibility for the final content of this work.

\section*{Ethics Statement}
This work involves no human subjects, no personally identifiable information,
and no new data collection; all datasets and pretrained potentials are publicly released, cited, and used under their licenses
(Appendix~\ref{app:baselines}). Our method reduces the number of density functional theory (DFT) calculations required at a controlled MLIP annotation risk, and makes downstream molecular screening more accessible to groups without large computing allocations. The main limitations are: Proposition~\ref{thm:crc} holds in expectation and only under
Assumption~\ref{ass:exch}, so a cutoff calibrated on one pool carries no guarantee on configurations from a different distribution, and it controls average risk rather than per-configuration correctness. The method may also be less reliable on datasets with extreme MLIP residuals, since the per-configuration loss is capped at $B_{\mathrm{cfg}}$. We state these limitations in Appendix~\ref{app:limitations}. We are aware of no other foreseeable harms, conflicts of interest, or fairness concerns.

\section*{Reproducibility Statement}
All experiments use public datasets and pretrained potentials, with sources, versions, and preprocessing documented in Appendix~\ref{app:baselines}. The structural descriptor is specified in full in Appendix~\ref{app:structural-vectors}; the error predictor architecture, training objective, and hyperparameters appear in Section~\ref{sec:error_predictor} and Appendix~\ref{app:baselines}; and the calibration procedure appears in Section~\ref{sec:crc}, with its proof in Appendix~\ref{app:proof}. The data splits, predictor seeds, and the 20 calibration/test reallocations are described in Appendix~\ref{app:baselines}. Code for the selector, the calibration procedure, and all scripts producing the reported tables and figures will be made publicly available.

\bibliography{references}

\appendix
\input{texts/appendix/app_related_work}
\input{texts/appendix/app_feature}
\input{texts/appendix/app_experiments}

\end{document}

%% file: tables/values.tex
\newcommand{\OursGain}{0.81}
\newcommand{\OursGainThree}{0.808}
\newcommand{\OursDFT}{57.8}
\newcommand{\OursRho}{0.289}
\newcommand{\RawGain}{0.50}

\newcommand{\RawDFT}{61.9}

\newcommand{\SupGain}{0.34}

\newcommand{\SupDFT}{67.5}
\newcommand{\SupRho}{0.251}
\newcommand{\RandDFT}{71.6}
\newcommand{\OracleDFT}{54.4}

\newcommand{\MAEReductionMin}{36.4}
\newcommand{\RiskMax}{1.333}
\newcommand{\RiskMeanMax}{1.0359}
\newcommand{\TruncationRateMax}{0.159}
\newcommand{\QCMLMassRemoved}{97.116}
\newcommand{\RetainedForceMax}{50.300}
\newcommand{\SupWholePoolRepeats}{14}

\newcommand{\MatchedRawSaving}{3.0}
\newcommand{\MatchedRhoWinsRaw}{6}
\newcommand{\MatchedRawMAEReduction}{2.0}
\newcommand{\MatchedMAEWinsRaw}{4}
\newcommand{\MatchedRawSuccess}{0.7}
\newcommand{\MatchedSuccessWinsRaw}{5}
\newcommand{\MatchedSupSaving}{4.0}

\newcommand{\MatchedSupMAEReduction}{9.7}
\newcommand{\MatchedMAEWinsSup}{5}
\newcommand{\MatchedSupSuccess}{1.5}
\newcommand{\MatchedSuccessWinsSup}{6}

\newcommand{\KnnOneGain}{0.31}
\newcommand{\KnnFiveGain}{0.33}
\newcommand{\NativeForceGain}{0.23}

%% file: tables/values_restored.tex
\newcommand{\PhysAlignedDFT}{55.0}

\newcommand{\TransferMeanG}{0.17}

\newcommand{\TransferInDomainG}{0.80}

\newcommand{\TransferPositive}{12}
\newcommand{\TransferMinG}{-0.09}
\newcommand{\TransferMaxG}{0.36}

\newcommand{\RawKTwoMin}{0.445}
\newcommand{\RawKTwoMax}{0.494}

\newcommand{\SupKTwoMin}{0.244}
\newcommand{\SupKTwoMax}{0.509}

%% file: texts/abstract.tex
Automated materials discovery requires evaluating large pools of candidate
configurations, with density functional theory (DFT) as a major throughput bottleneck.
Pretrained machine-learned interatomic potentials (MLIPs) approximate DFT cheaply, but their errors vary by orders of magnitude across configurations, making it difficult to decide when to trust the MLIP predictions and when to fall back to DFT.
We frame this decision as a risk-controlled selection problem and solve it within a conformal risk control (CRC) framework.
Each configuration is mapped from atomic
numbers and Cartesian coordinates to a fixed geometric descriptor, which a lightweight
predictor uses to estimate MLIP--DFT disagreement score in energy, average force, and maximum force.
CRC then calibrates this score into a selection threshold for DFT labeling such that the MLIP error on the unselected configurations, which keep their MLIP labels, stays below a
user-chosen fraction $\alpha$ of the expected error.
Under an identical guarantee across eight dataset–MLIP pairs, our predictor routes 7\% and 14\% fewer configurations to DFT at $\alpha = 0.3$ than committee baselines. Because committee disagreement overlooks errors the MLIPs share, our gains are largest when the DFT labels and MLIP training data differ in DFT functional and basis set, and narrow when the two agree. Without model access, the structure-only predictor also ranks errors as good as target MLIP’s latent representation and outperforms widely used descriptors like Smooth Overlap of Atomic Positions (SOAP) and Atom-Centered Symmetry Functions (ACSF).

%% file: texts/intro.tex

Density functional theory (DFT) provides accurate energies and atomic forces from electronic structure and is a central computational tool across chemistry and materials science \citep{kohn1965self}. However, conventional DFT scales approximately as $O(N^3)$ with system size, where $N$ denotes the number of atoms, limiting the number of atomic configurations that can be evaluated in large-scale applications such as molecular dynamics and high-throughput screening \citep{bowler2012on,curtarolo2013highthroughput}. Machine-learned interatomic potentials (MLIPs) address this bottleneck by approximating DFT energies and forces at a small fraction of the computational cost, typically with linear scaling in $N$ for local architectures \citep{behler2007generalized,unke2021machine,batatia2022mace}. Larger training datasets and increased model capacity have substantially improved the accuracy and generality of MLIPs, making pretrained models practical surrogates for DFT across broad regions of chemical space \citep{batatia2025foundation,wood2025uma}. However, no pretrained MLIP is uniformly accurate, and prediction errors can vary substantially across molecules, geometries, and chemical environments \citep{fu2023forces,deng2025systematic,perez2025uncertainty}.

This creates a basic decision problem whenever an MLIP should be used to accelerate a DFT workflow. The main question then becomes: Given a large pool of molecular configurations, \emph{which configurations can safely be evaluated with the MLIP and which should be sent to DFT?} Evaluating every configuration with DFT increases the computational cost, while accepting every MLIP prediction results in configurations with large energy or force errors. A natural strategy is to estimate the error for each configuration and send only the unreliable configurations to DFT \citep{jinnouchi2019onthefly,liu2026heterogeneous}. The most common estimate is the disagreement within a committee of MLIPs \citep{smith2018less,schran2020committee}. However, disagreement only flags errors on which the models differ, and errors that all models share remain invisible. The challenge is that the true error is unavailable without performing the expensive DFT calculation \citep{perez2025uncertainty}. Moreover, an error predictor that ranks configurations well does not by itself provide a finite-sample guarantee on the error remaining among configurations that are not evaluated with DFT \citep{angelopoulos2024conformal}. Predicting error is not the same as knowing where to stop trusting the MLIP.

\begin{figure}[t]
\centering
\includegraphics[width=\linewidth]{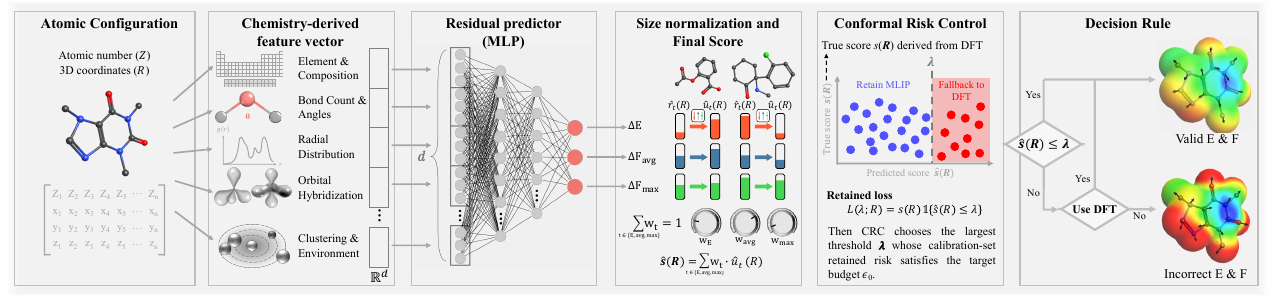}
\caption{\textbf{Overview of the MLIP--DFT selection framework.}
A molecular configuration $R$ (caffeine in this example) is converted into a chemistry-derived structural vector and passes through a lightweight predictor to estimate energy and force residuals. These estimates are normalized and combined into a predicted error score $\hat s(R)$. Conformal risk control then calibrates a threshold that retains reliable MLIP predictions and sends higher-risk configurations to DFT. The score and the capped loss are fixed before calibration.}
\label{fig:overview}
\end{figure}

In this work, we treat MLIP--DFT routing as risk-controlled selection for a frozen pretrained MLIP and solve it using conformal risk control (CRC)  \citep{bates2021distribution,angelopoulos2024conformal} (Figure~\ref{fig:overview}). The method consists of two steps. First, a lightweight predictor maps a structural representation rooted in analytical-chemistry consensus to the expected MLIP--DFT error in three channels: energy, average force, and maximum force. The channels are combined into a single predicted error score. Second, CRC uses a labeled calibration set to turn the score into a threshold, and every configuration scoring above this threshold is sent to DFT for labeling. Note the exchangeable unit of calibration is a \emph{molecule identity} (the set of all sampled geometries of one molecule). Routing decisions, however, are made for individual \emph{configurations} that potentially share the same molecule identity. Calibration controls validity. Under the exchangeability of molecule identities and for any error score fixed before calibration, the expected error left in the MLIP predictions of a test identity is at most a user-chosen fraction $\alpha$ of the average per-identity error on the fitting set. The quality of the error score predictor, or ranker, controls the efficiency, that is, how many DFT calls are needed to meet this budget, while CRC decides the score threshold that validates the risk control. 

Our \textbf{contributions} are as follows: (i) \textbf{Risk-controlled routing.} We cast MLIP--DFT routing as conformal risk control over molecular identities, calibrating a threshold for any given MLIP--DFT error score, so that the configurations scoring above it are sent to DFT while the MLIP error on the rest stays within a user-set budget. Score quality affects the number of DFT calls needed, not the guarantee (Section~\ref{sec:crc}). (ii) \textbf{A model-agnostic error ranker.} We propose a lightweight error predictor built on a structure-only, chemistry-informed descriptor. It ranks MLIP errors about as well as the target MLIP's own latent representation (Sections~\ref{sec:structural_vector}--\ref{sec:error_predictor}). (iii) \textbf{A guarantee-matched comparison of DFT cost.} At the same level of risk control, we compare  different error scores by the number of DFT calls they require across eight dataset--MLIP pairs. Our structure-based error score predictor requires fewer DFT calls than committee-based baselines. The gain is largest when the user's DFT reference differs from the MLIPs' training reference, and the gap closes when they match. We also show that partial DFT replacement can break error cancellation in energy differences (Section~\ref{sec:exp:main}, Appendix~\ref{app:cancellation}).

\paragraph{Related work.}
The closest precedents are \citet{liu2026heterogeneous}, who send structures to DFT using an empirically chosen threshold on ensemble uncertainty, and Conformal Arbitrage \citep{overman2025conformal}, which also calibrates a deferral threshold with CRC. We differ by using a separately trained error predictor and controlling a molecular identity-level loss. Appendix~\ref{app:related_work} gives a full review of the related literature.

%% file: texts/setup.tex
\subsection{Setup and notation}
Let $R$ denote a configuration with $N=N(R)$ atoms, atomic numbers $Z_j(R)$ and
positions $\mathbf x_j(R)\in\R^3$, $j=1,\ldots,N$. DFT assigns $R$ an energy $e^*(R)\in\R$ and per-atom forces $\mathbf f^*_j(R)\in\R^3$; a pretrained MLIP outputs predictions $\hat e(R)$ and $\hat{\mathbf f}_j(R)$ cheaply. The MLIP leaves residuals,
\begin{equation}\label{eq:residuals}
    \delta(R) = e^*(R)-\hat e(R)\in\R,
    \qquad
    \mathbf g_j(R) = \mathbf f^*_j(R)-\hat{\mathbf f}_j(R)\in\R^{3},
    \quad j=1,\ldots,N .
\end{equation}
MLIP and DFT labels do not share an isolated-atom energy convention, so we work with the aligned energy residual $\Delta E(R)=\delta(R)-\phi(R)^\top\hat a$, where $\phi(R)$ holds element counts and an intercept, and $\hat a$ contains the composition coefficients, fitted on $\Dfit$ and frozen before calibration (Appendix~\ref{app:baselines}). Because $\phi(R)^\top\hat a$ depends only on composition, the alignment leaves force residuals and energy differences between configurations of the same composition unchanged. Here, DFT is the reference the user trusts, which may use a different functional or basis set than the MLIP's training reference; the residual therefore also captures this systematic difference.
\paragraph{The routing problem.} Running DFT on a configuration drives its residual to zero, but costs far more than an MLIP forward pass. Given a frozen MLIP and an unlabeled pool of configurations, our goal is to send only a small subset of uncertain configurations to DFT and to accept MLIP predictions for the rest. However, a problem arises: deciding whether a configuration's residual is large enough to be sent to DFT requires the very DFT calculation we are trying to avoid. We therefore need a label-free proxy for the residual, together with a principled rule for how large that proxy must be before a DFT calculation is worth its cost.

\paragraph{Data splits.} We use three disjoint sets. $\Dfit$ and $\Dcal$ are DFT-labeled molecule identities that the MLIP was not trained on. We use $\Dfit$ to fit the MLIP error score predictor (Section~\ref{sec:error_predictor}) and $\Dcal$ to calibrate the selection threshold
(Section~\ref{sec:crc}). $\Dtest$ is the unlabeled test pool where the selection rule is applied. All configurations of the same molecule identity are assigned to the same split (Section~\ref{sec:crc}).

Our method proceeds in two steps. First, we use the DFT-labeled set $\Dfit$ to define a single error score that combines the MLIP energy residual and force residuals, and train a lightweight predictor $\hat s$ to estimate that error score from structural information alone; every configuration in $\Dtest$ then receives an estimated error score. Second, we use the error scores on $\Dcal$ to calibrate a risk-controlling threshold, and every configuration in $\Dtest$ scoring above the threshold is sent to DFT (Section~\ref{sec:crc}). Sections~\ref{sec:structural_vector} and~\ref{sec:error_predictor} describe the two halves of the first step, and Section~\ref{sec:crc} explains the conformal threshold and its risk-control guarantee.

%% file: texts/error_predictor.tex
\subsection{Structural vector}
\label{sec:structural_vector}

We define a structural vector constructed using only atomic numbers $Z_j$ and Cartesian coordinates $\mathbf x_j$. No other features, including DFT quantities, learned embeddings, or MLIP features, enter its construction.
The feature families are motivated by structural concepts from physical and analytical chemistry \citep{musil2021physics} and by established fixed-form descriptors such as atom-centered symmetry functions, the Coulomb matrix, and SOAP \citep{behler2011atom,rupp2012fast,bartok2013representing}. They encode composition, local coordination, pair-distance distortion, short-range crowding, angular distortion, radial structure, and elemental composition of neighboring coordination shells. These are geometric variables that chemists routinely use to reason about molecular structure from nuclear magnetic resonance, Fourier-transform infrared spectroscopy, and Raman spectroscopy \citep{geue2022structural,stavitski2010infrared}.

The descriptor concatenates nine fixed feature families into $350$ dimensions (Table~\ref{tab:vector-components} in Appendix~\ref{app:structural-vectors}). Coarse, discrete families count elements, element-resolved coordination states, and compressed, nominal, or stretched element-pair contacts. Continuous families then resolve what the counts hide: atomic-number-weighted short contacts, the few worst local environments, a smooth radial distribution, unweighted inverse-distance proximity at several powers, soft crowding across four distance scales, and the elemental identity of each atom's neighborhood. Exact definitions are also given in Appendix~\ref{app:structural-vectors}.

\subsection{A joint energy--force error predictor}
\label{sec:error_predictor}

Using the structural vector of Section~\ref{sec:structural_vector} as the input, we train a lightweight error predictor that estimates how accurate the MLIP energy and force prediction is. For a configuration $R$, the predictor estimates three targets: energy residual, average force residual, and maximum force residual,
\begin{equation}\label{eq:three_targets}
u_E(R)=|\Delta E(R)|, \qquad u_{\mathrm{avg}}(R)=\frac{1}{N}\sum_{j=1}^{N}\|\mathbf g_j(R)\|_2, \qquad u_{\max}(R)=\max_j\|\mathbf g_j(R)\|_2, \end{equation}
where $\Delta E(R)$ and $\mathbf g_j(R)$ are the aligned energy and per-atom force residuals of \eqref{eq:residuals}. The three targets capture different failure modes, each of which matters in practical molecular dynamics (MD) \citep{fu2023forces}. $u_E$ detects a wrong total energy, $u_{\mathrm{avg}}$ detects forces that are broadly inaccurate and mislead the trajectory, and $u_{\max}$ detects a single badly mispredicted atom that crashes the simulation. Because the three residual targets scale differently with atom count $N$, each is divided by a size-dependent scale $\hat c_t N^{\hat\beta_t}$ fitted on $\Dfit$ and then frozen, for $t\in\{E,\mathrm{avg},\max\}$  (detailed in Appendix~\ref{app:baselines}). We write $u_t$ for the normalized targets and $\hat u_t$ for their predictions.
The normalized residuals are combined into a single error score:
\begin{equation}
\label{eq:structural_score}
\hat s(R)=\sum_t w_t\hat u_t(R),
\qquad
s(R)=\sum_t w_tu_t(R),
\qquad
w_t\ge0,
\qquad
\sum_t w_t=1,
\end{equation}
where $s(R)$ denotes the oracle error score. The weights are fixed before calibration and specify the relative
importance of energy, average-force, and maximum-force errors for the intended
application---e.g.\ a user who cares more about force accuracy than energy accuracy can
set $w$ accordingly. All experiments in this paper use equal weights $w_t=1/3$.

%% file: texts/crc.tex

\subsection{Configuration selection by conformal risk control}\label{sec:crc}

The error predictor $\hat s$ can rank configurations but carries no guarantee about the true error of the unselected configurations---even a perfect ranking would not say where to stop, since that is
a decision about tolerable MLIP error against DFT cost, not a property of the predictor.
CRC sidesteps this: the user instead chooses $\alpha$, the fraction of the mean per-identity
MLIP error they are willing to tolerate, and calibration returns the threshold $\hat\lambda$
on $\hat s$ that achieves it. This exposes an explicit tradeoff between \emph{retained MLIP error} and \emph{the number of DFT
calls}---tolerating more error means fewer calls. Sweeping $\alpha$ traces this cost--accuracy
curve (Figure~\ref{fig:main}, middle row), and the practitioner picks an operating point on it
rather than guessing one in advance.

We follow the conformal risk control (CRC) framework of
\citet{angelopoulos2024conformal}, with one adaptation: the exchangeable unit is a
molecule identity, not a configuration \citep{dunn2023distribution}. We write $g$ for an identity and $\mathcal R_g$ for its set of configurations, and assign whole identities to splits
$\Dfit, \Dcal, \Dtest$ uniformly at random. When a dataset does not record the parent identity of a configuration, that configuration forms its own group (see Appendix~\ref{app:baselines}). We make the following exchangeability assumption:

\begin{assumption}[Exchangeability of molecule identities]\label{ass:exch}
The calibration identities $\Dcal=\{g_i\}_{i=1}^{n}$ and the test identities
$\Dtest=\{g_i\}_{i=n+1}^{n+m}$, each carrying its finite configuration set
$\mathcal{R}_{g_i}$ with finite scores, are exchangeable.
\end{assumption}

Selection, on the other hand, still happens one configuration at a time---a molecule identity may have some of its configurations sent to DFT and the rest left to the MLIP.
Let $s(R)$ be the true error score of Section~\ref{sec:error_predictor} and $\hat s(R)$
its estimate. CRC requires a bounded loss
\citep{bates2021distribution,angelopoulos2024conformal}, but $s$ is commonly heavy-tailed, so we
fix a truncation level $B_{\mathrm{cfg}}>0$ and set
\begin{equation}\label{eq:ell}
    \ell(R)=\min\{s(R),\,B_{\mathrm{cfg}}\}.
\end{equation}
$B_{\mathrm{cfg}}$ is a data-driven tolerance: the level past which a
configuration's error is unacceptable and its exact size no longer matters. Since the
distribution of error scores is skewed on several of our datasets, we set
$B_{\mathrm{cfg}}$ to the $99.9\%$ quantile of $\hat s$ on $\Dfit$ for robustness, rather than
to its maximum. At a threshold $\lambda$, we send to DFT every test configuration whose predicted
error score exceeds $\lambda$,
\begin{equation}\label{eq:selrule}
    S_\lambda=\{R\in\cT:\ \hat s(R)>\lambda\},
\end{equation}
and define the loss of a molecule identity $g$ as the total truncated error of its configurations
that are not selected, capped at $B$,
\begin{equation}\label{eq:loss}
    L(\lambda;g)=\min\Bigl\{\sum_{R\in\mathcal R_g}\ell(R)\,\mathbf 1\{\hat s(R)\le\lambda\},\ B\Bigr\}.
\end{equation}
The loss combines the true truncated error $\ell(R)$ with the predicted error score
$\hat s(R)$, which enters only through the ordering it induces. As a function of $\lambda$,
$L(\lambda;g)$ is non-decreasing and right-continuous, and it is zero once $\lambda$ is
small enough that every configuration goes to DFT---the conditions required by
\citet{angelopoulos2024conformal} after reversing the sign of $\lambda$. We sum rather than average over $\mathcal R_g$ because an identity with more unselected configurations left to the MLIP carries proportionally more error exposure. 

CRC additionally requires an upper bound on the loss of
a single exchangeable unit. Capping at $B$ inside \eqref{eq:loss} supplies that bound by construction:
$\sup_{\lambda}L(\lambda;g)\ \le\ B$. The two constants play distinct roles: $B_{\mathrm{cfg}}$ bounds the loss of an
individual configuration, whereas $B$ bounds the loss of an entire identity.
Appendix~\ref{app:baselines} details how each is estimated on $\Dfit$. Our goal is to choose $\hat\lambda$ on a calibration set $\Dcal$ of $n$ identities so that
the expected loss of a test identity $g_{n+1}$ is at most a user-specified level $\epsilon_0$:
$\E\bigl[L(\hat\lambda;g_{n+1})\bigr]\leq\epsilon_0$.

\begin{definition}[Finite calibrated selection threshold]\label{def:cutoff}
Let $\Dcal=\{g_i\}_{i=1}^n$ contain $n$ identities, and define the finite extended
set of observed configuration scores
\begin{equation}\label{eq:grid}
\mathcal H_n=\{-\infty\}\cup
 \bigcup_{i=1}^n\{\hat s(R):R\in\mathcal R_{g_i}\}.
\end{equation}
Given the budget $\epsilon_0>0$, let
\begin{equation}\label{eq:feasible}
\mathcal F_n=\left\{\lambda\in\mathcal H_n:
 \frac{\sum_{i=1}^n L(\lambda;g_i)+B}{n+1}\le\epsilon_0\right\}.
\end{equation}
The threshold is
\begin{equation}\label{eq:cutoff}
\hat\lambda=
\begin{cases}
\max\mathcal F_n,&\mathcal F_n\ne\varnothing,\\
-\infty,&\mathcal F_n=\varnothing.
\end{cases}
\end{equation}
\end{definition}
The selection rule is a threshold on $\hat s$, so configurations with identical
scores are retained or rejected together and only distinct score values are
reachable as thresholds. The term $B/(n+1)$ is the finite-sample correction for the unseen test identity: if $\epsilon_0<B/(n+1)$, then $\hat\lambda=-\infty$, and the whole pool is sent to DFT---the rule is never
infeasible; it simply forgoes all savings. The calibrated $\hat\lambda$ then controls the risk both for a single test identity and for the test pool as a whole.

\begin{proposition}[Risk control per identity]\label{thm:crc}
Let Assumption~\ref{ass:exch} hold, let $\hat s$, $B_{\mathrm{cfg}}$ and $B$ be fixed given
$\Dfit$. Given the threshold $\hat\lambda$
\eqref{eq:cutoff}, for a test pool $\Dtest$ with $m=|\Dtest|$ and a test identity $g_{n+1}$,
\begin{equation}
     \E\bigl[L(\hat\lambda;g_{n+1})\bigr]\ \le\ \epsilon_0,
    \qquad
    \E\left[\sum_{g\in\Dtest}L(\hat\lambda;g)\right]
    \;=\;
    \sum_{g\in\Dtest}\E\bigl[L(\hat\lambda;g)\bigr]
    \;\le\; m\,\epsilon_0 .
\end{equation}
\end{proposition}
The proof follows that of \citet{angelopoulos2024conformal}, and we provide the full proof in
Appendix~\ref{app:proof}. 

We set $\epsilon_0=\alpha\,\bar\ell_{\mathrm{fit}}$, where $\bar\ell_{\mathrm{fit}}$ is
the mean loss per identity on $\Dfit$. Thus, in expectation, the total error carried by
the unselected configurations of a test identity, capped at $B$, is at most an $\alpha$ fraction of the mean per-identity error of the fitting population. Choosing $\alpha=0.1$ therefore leaves about a
tenth of the error mass in the pool and removes the rest by DFT. Because the bound is anchored to the fitting population rather than the test pool, this does not assert that the realized pool retains at most an $\alpha$ fraction of its own error; Section~\ref{sec:experiments} reports both quantities. The expectation in Proposition~\ref{thm:crc} is over the exchangeable calibration and test identities, with the fitting stage held fixed. Because $\bar\ell_{\mathrm{fit}}$ is fixed by the fitting stage, the bound holds for each $\alpha$ chosen before calibration. It is not a high-probability bound for a single calibration draw, and it is not simultaneous over several values of $\alpha$ (Remark~\ref{rem:ties}).

\begin{remark}[Role of the predictor]\label{rem:predictorrole}
Proposition~\ref{thm:crc} holds for \emph{any} $\hat s$ fixed given $\Dfit$, regardless of the predictor's quality. An
uninformative predictor gives $\E[\ell\mathbf 1\{\hat s\le\lambda\}]=
\E[\ell]\mathbb{P}(\hat s \le\lambda)$, so the budget forces
$\mathbb{P}(\hat s\le\lambda)\lesssim\epsilon_0/\E[\ell]$ and nearly the entire pool is sent to
DFT. Predictor quality therefore governs DFT cost, not the validity of risk control; the quantity we are interested in is how many DFT evaluations are saved under a fixed budget $\epsilon_0$. The guarantee also covers scores chosen before calibration from $\Dfit$ or prior knowledge, such as whether the user's DFT matches the MLIP's training reference.
\end{remark}

%% file: texts/experiments.tex

\begin{table}[t]
  \centering\small
  \setlength{\tabcolsep}{3pt}
  \resizebox{\linewidth}{!}{%
  \begin{tabular}{lrrrl}
  \toprule
  Dataset & \# configs. & \# identities & \# atoms (mean) & elements \\
  \midrule
  COMP6-GDB07to09-FULL \citep{smith2018less} & 36{,}000 & 1{,}500 & 7--27 (17.0)   & H, C, N, O \\
  MACE-OFF23-FULL \citep{eastman2023spice}      & 50{,}195 & 15{,}769 & 3--150 (30.9)  & H, C, N, O, F, P, S, Cl, Br, I \\
  QCML-6950-9 \citep{ganscha2025qcml}          & 62{,}550 & 6{,}950  & 2--23 (11.0)   & H, C, N, O, F, P, S, Cl, Br, I \\
  \bottomrule
  \end{tabular}%
  }
  \caption{\textbf{Evaluation datasets.} A \emph{configuration} is a geometry and a \emph{molecule identity} is the exchangeable unit; configurations per identity differ by nearly an order of magnitude across datasets. MACE-OFF23 contains $228$ single-configuration groups without a recorded parent identity.}
  \label{tab:datasets}
\end{table}

\subsection{Setup}\label{sec:exp:setup}

We compare our score against four baselines that differ only in the predicted error score $\hat s$ used in \eqref{eq:loss}:
(i)~\textbf{Random}, where $\hat s$ is uniform and sets the reference cost;
(ii)~\textbf{Raw committee}, where $\hat s$ is the rank-combined disagreement among three pretrained MLIP architectures on predicted energy and forces, a standard uncertainty proxy for interatomic potentials \citep{smith2018less,schran2020committee,liu2026heterogeneous};
(iii)~\textbf{Supervised committee}, where $\hat s$ is a learned combination of the same three-model disagreement signals, produced by a residual-regression network trained on $\Dfit$; and
(iv)~\textbf{Oracle}, where $\hat s = s$ is the true error, a retrospective reference for the cost of perfect ranking rather than a proved lower bound.

All are wrapped in the identical CRC procedure of Section~\ref{sec:crc}, so all satisfy the same risk guarantee (further details in Appendix~\ref{app:baselines}). Note that each committee requires three pretrained-MLIP evaluations per configuration, whereas ours requires only one MLIP evaluation and the structural descriptor. We report these evaluation counts rather than wall-clock runtimes.

\paragraph{Datasets.} We evaluate our method on three molecular datasets that contain DFT labels (Table~\ref{tab:datasets}) and three pretrained MLIP models: \textbf{MACE-OFF23(M)} \citep{kovacs2025maceoff}, \textbf{AIMNet2} \citep{anstine2025aimnet2}, and \textbf{AceFF-2/TensorNet2} \citep{aceff2,simeon2023tensornet}. We do not evaluate MACE-OFF23(M) on MACE-OFF23-FULL, whose source it was trained on, leaving $8$ dataset--MLIP combinations, or \emph{cells}. Each MLIP is evaluated once per dataset; datasets and MLIPs use different electronic-structure references, so residuals include reference differences as well as model error (Appendix~\ref{app:provenance}).

\paragraph{Splits and calibration.}
We split each dataset by molecule identity into $\Dfit, \Dcal, \Dtest$ at $40/20/40$, so that all configurations of a given identity remain in the same split. The error predictor of Section~\ref{sec:error_predictor} is trained on $\Dfit$ and then frozen, along with energy alignment, size normalization, $B_{\mathrm{cfg}}$, $B$, and $\bar\ell_{\mathrm{fit}}$. For each cell, we draw $20$ random $\Dcal,\Dtest$ resamples and sweep $\alpha\in\{0.05,0.10,\ldots,0.50\}$, the target fraction of error retained. Proposition~\ref{thm:crc} converts into the calibration budget $\epsilon_0=\alpha\,\bar\ell_{\mathrm{fit}}$.

\paragraph{Fitting the predictor.}
Routing depends on the predictor only through the order it induces, so we ask it to rank well, not to predict accurately. The prediction $\hat u_t$ in Eq.~\ref{eq:structural_score} is the average of three small multilayer perceptrons (seeds $0$, $1$, $2$; two hidden layers of $256$ units) trained independently on $\Dfit$. Each network $b$ minimizes an $\ell_1$ residual loss on the three normalized channels plus a pairwise ranking loss on its own composite score $q_b(R)=\sum_t w_t\hat u_{b,t}(R)$. The ranking targets are the truncated error scores $\ell_i=\ell(R_i)$ of Eq.~\eqref{eq:ell} in Section~\ref{sec:crc}, the same quantity that CRC controls. For a mini-batch $\mathcal B$ with a uniformly permuted pairing $\pi$, let $\mathcal P=\{i\in\mathcal B:\ell_i\neq\ell_{\pi(i)}\}$,
\begin{equation}\label{eq:ranking-training}
\mathcal L_b=\frac{1}{|\mathcal B|}\sum_{i\in\mathcal B}\sum_t w_t\bigl|\hat u_{b,t}(R_i)-u_t(R_i)\bigr|
+\frac{\lambda_{\rm rank}}{|\mathcal P|}\sum_{i\in\mathcal P}
\log\!\Bigl[1+e^{-(q_b(R_i)-q_b(R_{\pi(i)}))\,\operatorname{sign}(\ell_i-\ell_{\pi(i)})}\Bigr].
\end{equation}
The ranking loss term pushes configurations with larger true loss toward larger scores; the regression term keeps the three channels interpretable. Architecture, seeds, and the value of $\lambda_{\rm rank}$ are chosen on a validation split internal to $\Dfit$ and frozen before calibration (Appendix~\ref{app:score-validation}).

\paragraph{Metrics.}
For each $\alpha$, we report three quantities:
(i)~the \textbf{DFT calls} $k(\alpha) = |S_{\hat\lambda(\alpha)}|$, the number of configurations routed to DFT at that $\alpha$, reported as a percentage of the test pool;
(ii)~the \textbf{risk ratio} $r(\alpha) = \sum_{g \in \Dtest} L(\hat\lambda(\alpha); g) \, / \, (|\Dtest| \, \epsilon_0)$, which measures how close the realized loss comes to the calibration budget and satisfies $\mathbb{E}[r(\alpha)] \le 1$ by Proposition~\ref{thm:crc}; and
(iii)~the \textbf{realized retained loss} $\rho(\alpha) = \sum_{R \in \cT \setminus S_{\hat\lambda(\alpha)}} \ell(R) \, / \, \sum_{R \in \cT} \ell(R)$, the error left uncorrected relative to the pool's total error rather than to $\epsilon_0$.

Since calls weight all configurations equally, we also report the $N^3$-weighted DFT cost proxy $C_{N^3}(\alpha)=100\sum_{R\in S_{\hat\lambda(\alpha)}}N(R)^3\big/\sum_{R\in\cT}N(R)^3$ (a size-scaling proxy, not wall time).
We also measure the normalized gain,
\begin{equation}\label{eq:gain}
G=\frac{\mathrm{AUC}_{\mathrm{rand}}-\mathrm{AUC}_{\hat s}}
       {\mathrm{AUC}_{\mathrm{rand}}-\mathrm{AUC}_{\mathrm{oracle}}},
\qquad
\mathrm{AUC}_{\hat s}=\frac{1}{\alpha_{\max}-\alpha_{\min}}\int k(\alpha)\,d\alpha
\end{equation}
This represents how much of the random-to-oracle  headroom is captured by score $\hat s$. Note $G=0$ and
$G=1$ correspond to random and oracle, respectively, and $G<0$ means it is worse than random selection (anti-selective).

\subsection{Results}\label{sec:exp:main}

\begin{figure}[t]
\centering
\includegraphics[width=\linewidth]{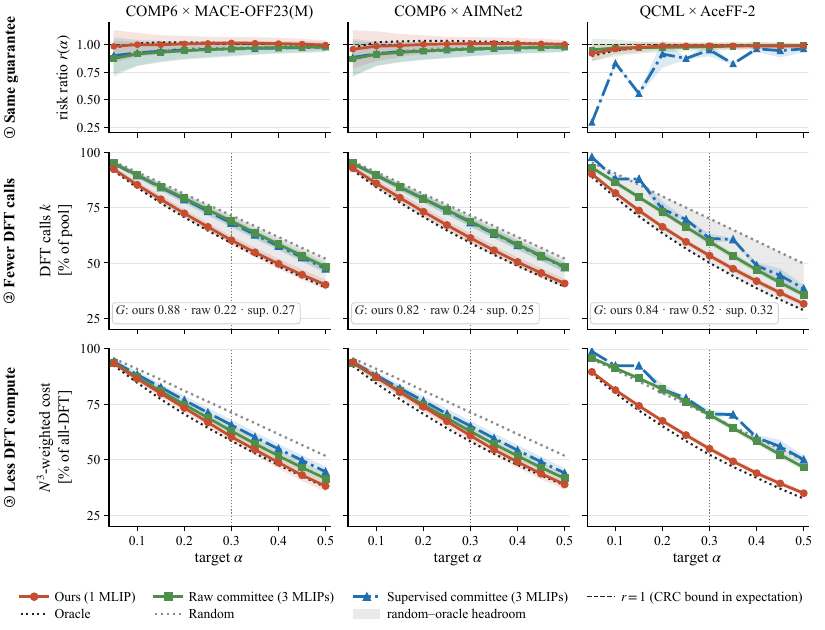}
\caption{\textbf{Every score meets the same guarantee; the DFT cost differs.} Three representative cells. \textbf{Top row}: risk ratio $r(\alpha)$ against the bound $r=1$. \textbf{Middle row}: DFT calls $k(\alpha)$, with the random-to-oracle headroom shaded and each cell's $G$ boxed. \textbf{Bottom row}: $N^3$-weighted DFT cost (\% of all-DFT), a size-scaling proxy, not wall time. Lines are means over 20 calibration/test reallocations, and bands are $\pm1$ SD for ours and the committees. Dotted verticals mark $\alpha=0.30$.}
\label{fig:main}
\end{figure}

\paragraph{Same guarantee, but lower DFT cost.} Every score passes through the identical CRC rule, so Proposition~\ref{thm:crc} gives all the same guarantee. With validity fixed by design, models are compared using DFT cost at certified risk. In the top row of Fig.~\ref{fig:main}, over all cells/scores/targets, the mean risk ratio is at most $\RiskMeanMax$ and the largest single draw is $\RiskMax$. Recall that the guarantee is on the expectation, not on each draw. The finite-sample correction $B/(n{+}1)$ ensures the metric is fair. Without it, the score violates the budget in roughly 52\% of calibration draws across $n=5$--100 identities. But with this correction, violations drop to 0.9\% at $n=10$, trading safety for higher DFT cost when calibration data is scarce (Appendix~\ref{app:guarantee}). Averaged uniformly over the eight cells (middle row), ours captures $G=\OursGain$ of the random-to-oracle headroom, compared to $\RawGain$ and $\SupGain$ for the raw and supervised committees, respectively. At $\alpha=0.30$, it routes $\OursDFT\%$ of the pool, against $\RawDFT\%$, $\SupDFT\%$, and $\RandDFT\%$ for raw committee, supervised committee, and random, respectively. Our DFT call was comparable with oracle ($\OracleDFT\%$). Under $N^3$ weighting (bottom row) and $\alpha\ge0.15$, ours is cheaper than the raw committee and random in all eight cells and than the supervised committee in seven of eight cells (Appendix~\ref{app:cost}). The realized retained fraction $\rho(\alpha)$ tracks $\alpha$ for every score (Fig.~\ref{fig:app-E2}, Appendix~\ref{app:guarantee}). At $\alpha=0.30$, ours retains $\rho=\OursRho$, closest to the target, while the supervised committee ($\SupRho$) spends extra DFT on error the user would have tolerated.

\begin{figure}[t]
\centering
\includegraphics[width=\linewidth]{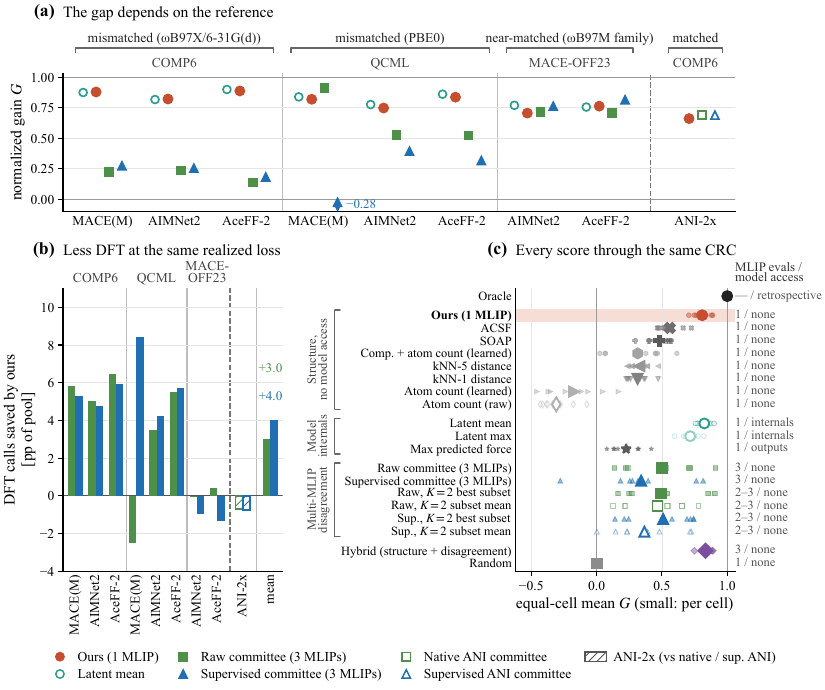}
\caption{\textbf{Structure beats disagreement where the MLIPs share error; the gap closes as references converge.} (a) Per-cell $G$. Brackets give the relation between dataset and MLIP training references. Right of the dashed line, ANI-2x $\times$ COMP6 is a different MLIP and population, compared with its native and supervised ANI committees. The supervised committee on QCML $\times$ MACE-OFF23(M) ($G=-0.28$) is drawn clipped. (b) DFT saved by ours at matched realized loss $\rho$ (committee minus ours, pp of pool), averaged over the common attained $\rho$ range; the mean excludes ANI-2x. (c) Equal-cell mean $G$ (small markers: cells) for every score under the same CRC rule; the right column gives MLIP evaluations per configuration and required model access.}
\label{fig:gap}
\end{figure}

\paragraph{Disagreement misses error that the MLIPs share.} A residual mixes model error with any mismatch between the MLIP's training reference and the dataset's training reference. In such cases, the three MLIPs share common failure modes. Because they agree on the wrong answer, disagreement-based metrics collectively miss the correct answer. By contrast, our structure-only score captures this shared error from $\Dfit$. Fig.~\ref{fig:gap}(a) shows that on COMP6 (mismatch between DFT--MLIP training references), ours reaches $G=0.82$--0.89 versus 0.14--0.27 for the committees. On QCML (slightly less of a mismatch), 0.75--0.84 versus 0.32--0.53. One exception is QCML $\times$ MACE-OFF23(M), where the raw committee wins (0.91), while the supervised one falls below random ($-0.28$). We also observed that a few catastrophic, model-specific failures carried nearly all the error (Gini 1.00, against $\le0.18$ on COMP6). In comparison, on MACE-OFF23 cells (high alignment between DFT--MLIP training references) the supervised committee leads (0.76 and 0.82 versus 0.71 and 0.76). If disagreement misses shared error, committees slightly recover once the training references match. ANI-2x was trained using COMP6's standard reference. There, its own committee reaches $G=0.69$ and ours reaches 0.66 (whether ANI-2x saw these configurations during training can not be established). The shared error is also not just the standard dispersion correction. A score built from D3(BJ) alone reaches only $G\le0.12$, and adding D3(BJ) to the reference changes our $G$ by at most 0.02 (Appendix~\ref{app:mechanism}), meaning that our score is not just relearning the dispersion correction. Finally, the committees are not weak by design. Two-member committees average $G\le0.51$, and committees built from the target's own model family never beat ours in the five cells where they exist (Appendix~\ref{app:baselines-strength}). At matched realized loss $\rho$, ours still saves $\MatchedRawSaving$ and $\MatchedSupSaving$ percentage points of DFT against the raw and supervised committees, positive in $\MatchedRhoWinsRaw$ of 8 cells each (Fig.~\ref{fig:gap}(b), Appendix~\ref{app:cost}). The reference relation is known before calibration, so the score can be chosen without touching the guarantee (Remark~\ref{rem:predictorrole}). If unsure whether the DFT matches the MLIPs' training reference, the hybrid score (our descriptor plus the three disagreement channels) is always a safe choice. It is at least as good as ours in all eight cells, at the cost of running three MLIPs. This pattern comes from comparing nine settings that are not from a controlled experiment, supporting this explanation. In particular, MACE-OFF23 cells have both a closer reference and larger molecules, so they cannot say which of the two helps the committees.

\paragraph{Structure recovers what model internals know.} In Fig.~\ref{fig:gap}(c), ours ($G=\OursGain$) is on par with the target MLIP's own representation of 0.82, which needs model access and is fragile to pooling (max pooling: 0.72). It also far exceeds generic descriptors frequently used in chemistry like Atom-Centered Symmetry Functions (ACSF, 0.56) and the Smooth Overlap of Atomic Positions (SOAP, 0.48). Only the latent mean and the hybrid (0.83) are marginally higher, and among scores needing no model internals, ours is the strongest. CRC sees a score only through its order, and across all scores $G$ rises with the test Spearman correlation between $\hat s$ and $s$. Three checks show where this ranking comes from (Appendix~\ref{app:diagnostics}). First, no single feature family is significantly essential over another. Removing any one of them lowers $G$ by at most 0.03. Second, more fitting data helps. $G$ rises from 0.54 with 2.5\% of identities to 0.81 with 40\%. Third, the pairwise ranking loss is not needed: weights of 0, 0.1, and 1 give nearly the same $G$ of 0.807, 0.806, and 0.808, respectively. As an extension beyond the guarantee, we were curious on what routing does to relative energies (energy differences between geometries of one molecule). At $\alpha=0.30$, ours cuts their MAE from the all-MLIP level by at least $\MAEReductionMin\%$ in every cell, yet picking each molecule's lowest-energy geometry succeeds less often than with no DFT on the three COMP6 cells. One molecule's MLIP errors tend to share sign and cancel in differences, and replacing some with DFT removes that cancellation (Appendix~\ref{app:physical-grid}). This is a property of partial DFT substitution rather than of any particular score, so a guarantee on energy and force errors does not by itself cover the relative energies. Because Proposition~\ref{thm:crc} applies to any bounded loss that decreases as the threshold tightens (Remark~\ref{rem:otherloss}), our framework can certify these objectives directly. The required DFT compute depends on the target: 79–97\% for relative-energy errors, but only 8–61\% for selection failure (Appendix~\ref{app:other-losses}).

%% file: texts/conclusion.tex
We cast MLIP--DFT routing as risk-controlled selection. A predictor ranks configurations by expected error, and conformal risk control decides where DFT is deployed. Under exchangeability of molecule identities, the expected capped error left to MLIP stays within a chosen fraction of the fitting-set mean for any score fixed before calibration. Ranking quality therefore changes only DFT cost, and scores can be compared on that cost alone. This comparison matters most when the user's DFT reference differs from the MLIPs' training reference. In this scenario, the MLIPs share much of their error and agree on the wrong answer. Averaged over eight dataset--MLIP cells, a structure-only score fitted on labeled residuals captures $\OursGain$ of the random-to-oracle headroom, against $\RawGain$ and $\SupGain$ for three-MLIP committees. It also matches the target MLIP's own latent features without accessing them. When the references do match, committees recover and demonstrate improved results. The choice of score should therefore follow the reference relation, which is known before calibration. Two results define where the method applies: (i) the fitting and calibration labels are amortized over repeated screening of a population, and (ii) partial DFT replacement can break error cancellation in relative energies, so these objectives should be certified directly, which the framework supports. We provide further limitations and broader impacts in Appendix~\ref{app:limitations} and ~\ref{app:broader-impacts}.

%% file: texts/appendix/app_related_work.tex
\section{Related Works}\label{app:related_work}

\paragraph{Configuration-level uncertainty for pretrained MLIPs.}
Pretrained MLIPs are trained on large and chemically diverse data \citep{kovacs2025maceoff,batatia2025foundation,wood2025uma}, yet they still fail on individual geometries and retain systematic errors at scale \citep{fu2023forces,deng2025systematic}. Configuration-level uncertainty is commonly estimated from committee disagreement \citep{peterson2017addressing,schran2020committee}, Bayesian or evidential models \citep{vandermause2020onthefly,xu2026evidential}, or distances in a learned latent space \citep{hu2022robust}. For large pretrained models, cheaper estimators avoid training full ensembles \citep{bilbrey2025uncertainty,bigi2024prediction,kellner2024uncertainty}. \citet{liu2026heterogeneous} combine a heterogeneous ensemble of universal potentials into one uncertainty metric and use an empirically chosen cutoff on it to decide which structures need DFT labels. Such estimates are often overconfident and must be calibrated against labeled errors before use \citep{perez2025uncertainty}. Raw and supervised committees represent this lineage as baselines. In our framework, however, the cutoff on any score is calibrated by a CRC framework rather than chosen empirically.

\paragraph{Post-hoc error and reliability prediction for trained MLIPs.}
There are several methods that share the idea of training a  predictor that estimates the error of a pretrained model. DEUP trains a secondary predictor that estimates out-of-sample error \citep{lahlou2023deup}, whereas $\Delta$-learning learns a signed correction to a cheap baseline \citep{ramakrishnan2015big}. For MLIPs, recent post-hoc methods draw on the target model's latent representation or training history \citep{vita2025ltau,huang2025pdrum,mehdi2026knowing}. P-DRUM is closest in target because it regresses energy and force residuals from the descriptors of a trained MACE model \citep{huang2025pdrum}. PROBE instead recasts energy reliability as selective classification on frozen per-atom representations \citep{mehdi2026knowing}. Our predictor uses a structure-only descriptor whose construction does not depend on the target MLIP (Section~\ref{sec:structural_vector}), and a separate predictor is fitted for each dataset--MLIP cell. It needs no model internals but forgoes the model-specific information that a latent representation may carry. Only its ordering enters the routing decision, and the threshold is calibrated separately (Section~\ref{sec:crc}).

\paragraph{Uncertainty-guided DFT calls, acquisition, and screening.}
Learn-on-the-fly molecular dynamics has long coupled quantum-mechanical calculations with a force model that is refitted or extended during the simulation \citep{csanyi2004learn,li2015molecular}. Modern on-the-fly schemes call DFT when a Bayesian error estimate is large and add the result to the training set \citep{jinnouchi2019onthefly,vandermause2020onthefly}. The DFT result can also replace the unreliable prediction at that step \citep{jinnouchi2019onthefly}. Active-learning pipelines similarly select configurations by committee disagreement or extrapolation grade and retrain on the new labels \citep{podryabinkin2017active,smith2018less,zhang2020dpgen}. Using the DFT value of a flagged configuration is therefore not specific to our setting. What differs is that our target MLIP is never retrained and each deployment pool is routed in a single pass. Our cutoff also carries a finite-sample guarantee on the expected capped retained loss, whereas these cutoffs are set by rule or tuned empirically. Screen-then-verify workflows use machine-learned models to pre-filter candidates for DFT, as in AdsorbML and the Matbench Discovery benchmark \citep{lan2023adsorbml,riebesell2025framework}. They select on predicted energy or stability rather than on predicted error. Our selection-success metric is analogous to the AdsorbML success rate (Appendix~\ref{app:physical-grid}).

\paragraph{Conformal calibration and risk control.}
Conformal prediction calibrates prediction sets or intervals to finite-sample coverage under exchangeability \citep{vovk2005algorithmic,angelopoulos2023gentle}. Risk-controlling prediction sets and Learn-then-Test control more general risks with high probability \citep{bates2021distribution,angelopoulos2025learn}, while CRC controls the expected value of a bounded monotone loss \citep{angelopoulos2024conformal}. Related calibration rules select trustworthy foundation-model outputs, filter augmented data, and select candidates with conformal p-values \citep{gui2024conformal,wu2025filtering,jin2023selection}. For MLIPs, \citet{hu2022robust} calibrate energy intervals with split conformal prediction, and \citet{ho2026flexible} learn a conformal-inspired scaling of per-atom force uncertainty without a formal coverage guarantee. Both calibrate the uncertainty of individual predictions, whereas our calibration sets a DFT-routing cutoff to control loss of an identity. For our guarantee to hold, we only need exchangeability across molecule identities because each identity contributes to a single bounded loss.

%% file: texts/appendix/app_feature.tex
\section{The 350-dimensional structural descriptor}\label{app:structural-vectors}
The descriptor is a deterministic function of atomic numbers $Z_j$ and Cartesian
coordinates $\mathbf x_j$. It uses no reference labels, MLIP predictions, learned
embeddings, or fitted parameters. Its nine feature families summarize composition,
coordination, distance distortion, close contacts, and local environments
(Table~\ref{tab:vector-components}). These are geometric proxies motivated by
chemical structure; they do not compute electronic structure.

\begin{table}[t]
\centering\small
\setlength{\tabcolsep}{4pt}
\renewcommand{\arraystretch}{1.08}
\begin{tabularx}{\linewidth}{lrlX}
\toprule
\textbf{Feature family} & \textbf{Dim.} & \textbf{Type} & \textbf{Description} \\
\midrule
$\mathrm{ElementCount}$ & 10 & Discrete & Global counts of the allowed elements. \\
$\mathrm{Coordination}$ & 52 & Discrete & Element-resolved proximity-graph coordination states. \\
$\mathrm{PairState}$ & 156 & Discrete & Element-pair contacts classified as compressed, near nominal separation, or elongated. \\
$\mathrm{ShortRange}$ & 21 & Continuous & Atomic-number-weighted severity of short interatomic contacts. \\
$\mathrm{Extreme}$ & 12 & Continuous & Most severe local proximity, strain, and angular-distortion environments. \\
$\mathrm{RBF}$ & 18 & Continuous & Smooth multiscale pair-distance channels with global and local pooling. \\
$\mathrm{InverseDistance}$ & 36 & Continuous & Element-independent geometric proximity across several inverse-distance scales. \\
$\mathrm{SoftCrowding}$ & 24 & Continuous & Smooth filling of local coordination shells over several radii. \\
$Z\mathrm{Potential}$ & 21 & Continuous & Neighbor-$Z$-weighted description of the local chemical environment. \\
\bottomrule
\end{tabularx}
\caption{\textbf{Feature families of the structural vector.} The canonical Coordination and PairState blocks contain $60$ and $165$ coordinates; a fixed compaction removes eight and nine of them, leaving $52$ and $156$. The descriptor contains the nine families and has $350$ dimensions.}
\label{tab:vector-components}
\end{table}

\subsection{Geometric quantities and summary conventions}
The fixed element order is H, C, N, O, F, P, S, Cl, Br, I, with atomic numbers
$[1,6,7,8,9,15,16,17,35,53]$. Covalent radii $\rho(Z)$ in \AA\ are
$[0.31,0.76,0.71,0.66,0.57,1.07,1.05,1.02,1.20,1.39]$
\citep{cordero2008covalent}; carbon uses the $sp^3$ value. Define
\[
r_{jk}=\|\mathbf x_j-\mathbf x_k\|_2,\qquad
q_{jk}=\frac{r_{jk}}{\rho(Z_j)+\rho(Z_k)},\qquad
\widetilde r_{jk}=\max(r_{jk},0.35\,\text{\AA}).
\]
The graph $G_0$ contains all pairs with $q_{jk}\le1.20$. A second graph $G$
uses the same candidate edges but restricts H and F to at most one connection,
retaining the lowest $(q_{jk},e)$ where $e$ is the edge identifier used to break
distance ties. Coordination and pair-state counts use $G_0$; local strain and
angular summaries use $G$.

For an atomwise or pairwise vector $x$, define the ordered summaries
\[
\begin{aligned}
\mathcal S_7(x)&=[\operatorname{sum},\operatorname{mean},\operatorname{std}_{\rm pop},
\max,Q_{0.90},\operatorname{rms},\operatorname{count}],\\
\mathcal U_6(x)&=[\operatorname{sum},\operatorname{mean},\operatorname{std}_{\rm pop},
\max,Q_{0.90},\operatorname{rms}].
\end{aligned}
\]
The operators $\operatorname{top1}$, $\operatorname{top2}$, and
$\operatorname{top3}$ sort values in descending order and pad with zeros if
fewer than three values exist. On empty edge or pair sets, all summaries and
fractions return zero. These conventions fix the dimension for small structures.

\subsection{Composition, coordination, and pair states: 218 coordinates}
ElementCount records ten element counts. For each element, Coordination counts
atoms with $G_0$ degrees $0,1,2,3,4,\ge5$, giving $60$ canonical coordinates.
For each of the $55$ unordered element pairs, PairState counts $G_0$ edges in
three regions: $q<0.95$, $0.95\le q\le1.05$, and $1.05<q\le1.20$, giving
$165$ canonical coordinates. These bins describe distances relative to
covalent-radius sums, without assigning bond order or bond energy.

Concatenate the $10+60+165=235$ canonical coordinates and remove the fixed
zero-based indices
\[
[15,16,22,33,38,39,40,46,133,154,172,173,174,181,182,183,184].
\]
Preserve the order of all remaining coordinates. Eight exclusions belong to
Coordination and nine to PairState, leaving $10+52+156=218$ inputs.

\subsection{Short-range and extreme local environments: 33 coordinates}
For powers $p=2,4,6$, define
\[
S_j^{(p)}=\sum_{k\ne j}\frac{Z_jZ_k}{\widetilde r_{jk}^{\,p}}.
\]
Append $\mathcal S_7(S^{(p)})$ for each power, giving $21$ ShortRange
coordinates; the count entry equals $N$. Higher powers emphasize closer
contacts. Extreme uses three atomwise quantities:
\[
R_j=S_j^{(2)},\qquad
W_j=\sum_{\substack{e\in G\\e\ni j}}
\exp\!\left[-\frac{(q_e-1)^2}{2(0.20)^2}\right](q_e-1)^2,
\]
\[
B_j=\sum_{\substack{k<\ell\\k,\ell\in\mathcal N_G(j)}}
|\sin(\theta_{kj\ell}-\theta_0)|,\qquad
\theta_0=\begin{cases}
\pi,&\deg_G(j)=2,\\
2\pi/3,&\deg_G(j)=3,\\
\arccos(-1/3),&\deg_G(j)\ge4.
\end{cases}
\]
Set $B_j=0$ below degree two. For each of $R,W,B$, append
$[\operatorname{top1},\operatorname{top2},\operatorname{top3},\operatorname{sum}]$,
giving $12$ coordinates. The angular reference is degree-based, not an orbital
hybridization assignment. Its non-monotone sine penalty can vanish for a
collapsed degree-two angle; the distance features still record close contacts.

\subsection{Radial and inverse-distance summaries: 54 coordinates}
For nine ordered centers
\[
(\mu_1,\ldots,\mu_9)=(0.568,0.728,0.888,1.048,1.208,1.368,1.528,1.688,1.848),
\]
define
\[
g_\mu(q)=\exp\!\left[-\tfrac12((q-\mu)/0.124)^2\right],\quad
P_\mu=\sum_{j<k}g_\mu(q_{jk}),\quad
M_\mu=\max_j\sum_{k\ne j}g_\mu(q_{jk}).
\]
The RBF block is $[P_{\mu_1},M_{\mu_1},\ldots,P_{\mu_9},M_{\mu_9}]$.
For each $p=1,\ldots,6$, InverseDistance appends
$\mathcal U_6(\{\widetilde r_{jk}^{-p}:j<k\})$, giving $36$ coordinates.
These unweighted proximity summaries distinguish distance effects from the
atomic-number weights in ShortRange.

\subsection{Soft crowding and neighboring elements: 45 coordinates}
For the ordered scale pairs
$(q_0,w)=(0.85,0.12),(1.00,0.16),(1.20,0.22),(1.50,0.30)$, define
\[
C_j^{(q_0,w)}=\sum_{k\ne j}\frac{1}{1+\exp[(q_{jk}-q_0)/w]}.
\]
At each scale append $[\operatorname{mean}C,\operatorname{std}_{\rm pop}C,
\min C,\max C,Q_{0.90}(C)]$. Then append, in order,
\[
\begin{aligned}
&\operatorname{mean}(C^{0.85,0.12}-C^{1.00,0.16}),\quad
\operatorname{mean}(C^{1.00,0.16}-C^{1.20,0.22}),\\
&\max(C^{0.85,0.12}-C^{1.20,0.22}),\quad
\operatorname{std}_{\rm pop}(C^{0.85,0.12}-C^{1.50,0.30}).
\end{aligned}
\]
These $24$ SoftCrowding coordinates use heuristic distance scales, not
validated physical coordination-shell boundaries. Finally, for $p=2,3,4$,
define $L_j^{(p)}=\sum_{k\ne j}Z_k/\widetilde r_{jk}^{\,p}$ and append
$\mathcal S_7(L^{(p)})$. The resulting $21$ $Z$Potential coordinates retain
neighbor element identity without multiplying by the central atomic number.
The complete dimension is $218+21+12+18+36+24+21=350$.

%% file: texts/appendix/app_experiments.tex

\section{Proof of Proposition~\ref{thm:crc}}\label{app:proof}

\begin{proof}[Proof of Proposition~\ref{thm:crc}]
Fix $\Dfit$, $\hat s$, target normalization, weights $w_t$, bounding constants $B_{\mathrm{cfg}}$ and $B$, and $\epsilon_0$. For each $\lambda$, the map $g\mapsto L(\lambda;g)$ is a fixed measurable function of a molecule and Assumption~\ref{ass:exch} makes
$\{L(\lambda;g_i)\}_{i\in\Dcal}\cup\{L(\lambda;g_{n+1})\}$ exchangeable. By construction, $L(\lambda; g_i)$ is a non-decreasing function of $\lambda$, $0\le L(\lambda; g_i)\le B$, and $L(-\infty; g_i)=0$.

Let $$\mathcal H_{n+1}=\mathcal H_n\cup\{\hat s(R):R\in\mathcal R_{g_{n+1}}\}$$ where $\mathcal H_n$ follows Eq.~\ref{eq:grid} and
$$
\lambda^\star=\max\Bigl\{\lambda\in\mathcal H_{n+1}:
 \tfrac{1}{n+1}\textstyle\sum_{i=1}^{n+1}L(\lambda;g_i)\le\epsilon_0\Bigr\},
$$
which denotes $\lambda$ in the oracle case when $L(\lambda;g_{n+1})$ is known. The set is non-empty since the condition always holds for $\lambda=-\infty$ so $\lambda^*$ is well-defined.

For any $\lambda\in\mathcal F_n$ we have $\lambda\in\mathcal H_{n+1}$ and
$$
\frac{1}{n+1}\sum_{i=1}^{n+1}L_i(\lambda)
\le\frac{\sum_{i=1}^{n}L_i(\lambda)+B}{n+1}\le\epsilon_0 ,
$$
so $\lambda\le\lambda^\star$; taking the maximum, $\hat\lambda\le\lambda^\star$
for $\hat\lambda$ defined in Eq.~\ref{eq:cutoff} (trivially so if $\mathcal F_n=\varnothing$).

Both $\mathcal H_{n+1}$ and $\sum_{i=1}^{n+1}L_i$ are invariant under permutations of $g_1,\dots,g_{n+1}$, so $\lambda^\star$ is a symmetric function of them. Assumption~\ref{ass:exch} then makes $\E[L(\lambda^\star;g_i)]$ constant in $i$, whence
$$
\E\bigl[L(\lambda^\star, g_{n+1})\bigr]
=\E\Bigl[\tfrac{1}{n+1}\textstyle\sum_{i=1}^{n+1}L(\lambda^\star, g_{i})\Bigr]
\le\epsilon_0 .
$$
Since $\hat\lambda\le\lambda^\star$ and $L_{n+1}$ is non-decreasing,
$L(\hat\lambda;g_{n+1})\le L(\lambda^\star;g_{n+1})$ point-wise, giving
$\E[L(\hat\lambda;g_{n+1})]\le\epsilon_0$. Each $g\in\Dtest$ is exchangeable
with $\Dcal$, so this bound applies and linearity of expectation yields
$\E[\sum_{g\in\Dtest}L(\hat\lambda;g)]\le m\epsilon_0$, with no independence required among test molecules.
\end{proof}

\begin{remark}[Ties, fallback, and scope]\label{rem:ties}
The argument is finite and needs neither continuity of $\hat s$ nor absence of
ties. When no finite candidate is feasible but
$\epsilon_0\ge B/(n+1)$, $\lambda=-\infty$ still satisfies the
criterion; when $\epsilon_0<B/(n+1)$, the criterion is infeasible, and the rule in Eq.~\ref{eq:cutoff} still sends every configuration to DFT. Our method controls \emph{expected} capped
identity-level retained loss for an $\alpha$ fixed before calibration and it is worth noting that it does not guarantee $\rho(\alpha)\le\alpha$ for a particular pool and does not bound uncapped physical
energy or force error.
\end{remark}

\begin{remark}[Other bounded losses]\label{rem:otherloss}
The proof only requires that $L(\lambda;g)$ is non-decreasing in $\lambda$, bounded by $B$, zero at $\lambda=-\infty$, and fixed given $\Dfit$. Any per-configuration loss $\ell'$ with these properties can replace $\ell$, and any aggregation over molecules that preserves monotonicity and the bound can replace the capped sum.
\end{remark}

\section{Datasets, experimental protocol, and baselines}\label{app:baselines}

We first specify what is fitted, what is frozen, and what calibration is allowed to use. All primary scores are calibrated against the same loss, with the same splits and target grid. Table~\ref{tab:datasets} summarizes the three evaluation datasets.

\paragraph{Splits and resampling.}
Each dataset is split by its stored molecule identity into $\Dfit,\Dcal,\Dtest$ at $40/20/40$.
Every fitting-derived quantity and the score predictor are estimated on $\Dfit$ once and
frozen. Calibration and test identities are reallocated in $20$ random resamples per
combination, with $\alpha\in\{0.05,0.10,\ldots,0.50\}$. 
On COMP6, each resample calibrates on $300$ identities ($7{,}200$ configurations) and
evaluates on $600$ ($14{,}400$); on QCML, it calibrates on $1{,}390$ ($12{,}510$) and evaluates on $2{,}780$
($25{,}020$). On MACE-OFF23, $6{,}308$ fitting groups contain $19{,}654$ configurations
with $1$--$229$ configurations per group; calibration and test sets contain $3{,}154$ and $6{,}307$
groups, with test-pool size $20{,}372.8\pm333.5$ configurations across resamples.
The MACE-OFF23 population has $15{,}769$ groups: $15{,}541$ known parent identities and $228$ configurations with unavailable parent identity, each kept as a singleton group. These singletons are not known to be distinct molecules; the exchangeability assumption is stated for the groups as executed. Total atom counts are $612{,}360$ for COMP6, $1{,}551{,}806$ for MACE-OFF23, and $687{,}204$ for QCML.

\paragraph{Predictor fitting.}
Each of the three scorers has hidden widths $(256,256)$, SiLU activations, dropout $0.1$, and three Softplus outputs. Training minimizes \eqref{eq:ranking-training} with AdamW, learning rate $10^{-3}$, weight decay $10^{-4}$, batch size $256$, and at most $300$ epochs. Checkpoints use the residual validation $L_1$ loss with patience $25$ and minimum improvement $10^{-5}$; the ranking loss is not used for early stopping. Feature centering, scaling, and removal of constant coordinates use only the internal training rows, while the residual normalization and loss transforms use the full $\Dfit$. Scorers are trained independently with seeds $0,1,2$. The arithmetic mean of the three scores is used directly. The same frozen checkpoints serve all $20$ calibration/test allocations, and there is no refit after selection.
 
\paragraph{Fitting set-derived quantities.}
The following are estimated on $\Dfit$ and then frozen at calibration: the per-element energy offsets $\hat a$ that place the MLIP and DFT energies on a common baseline; the size exponents $\hat \beta_t$ and normalizing factors $\hat c_t$ for the three channels $t\in\{E,\mathrm{avg},\max\}$, which together define the error score $s$; the truncation level $B_{\mathrm{cfg}}$, the $99.9\%$ quantile of $s$ over $\Dfit$, gives the loss $\ell=\min(s,B_{\mathrm{cfg}})$ of Eq.~\eqref{eq:ell}.
Because molecule identity is the exchangeable unit, the remaining two quantities are calculated over unique identities in $\Dfit$: the bound $B=\max_{g\in\mathcal I_{\mathrm{fit}}}\ \sum_{i\in g}\ell_i$ used in Eq.~\eqref{eq:feasible}, and the mean per-identity loss $\bar\ell_{\mathrm{fit}}=|\mathcal I_{\mathrm{fit}}|^{-1}\sum_{g\in\mathcal I_{\mathrm{fit}}}\sum_{i\in g}\ell_i$ setting the budget
$\epsilon_0=\alpha\,\bar\ell_{\mathrm{fit}}$. Here, $\mathcal I_{\mathrm{fit}}$ denotes the set of molecule identities of $\Dfit$.
Beyond these, the committees fit inter-model energy offsets to align the scales of different model predictions, and our model trains the score predictor $\hat s$, both from $\Dfit$. Random estimates nothing---$\hat s$ is a uniform draw---and oracle estimates nothing further, since $\hat s=s$. $\Dcal$ is spent entirely on choosing the threshold $\hat\lambda$ via Eq.~\eqref{eq:cutoff}.

\paragraph{Per-element energy alignment.}
A frozen public MLIP checkpoint and a reference DFT dataset do not share an isolated-atom energy convention. Their total energies therefore differ by a fixed amount per atom of each element, and that constant is far larger than the model error we want to measure.
The offset of a molecule is the sum of the per-element constants over its atoms, so it is exactly linear in the element counts:
\begin{equation}\label{eq:refalign}
    e_{\mathrm{true}}(R)-e_{\mathrm{pred}}(R)=\phi(R)^\top a+\varepsilon(R),
\end{equation}
where $\phi(R)\in\R^{K+1}$ holds the counts of the elements together with an intercept, and $a$ collects the
per-element offsets $a_z$ and that intercept. Regressing the energy difference on $\phi$ over $\Dfit$ recovers $\hat a$, and $\Delta E(R)=e_{\mathrm{true}}(R)-e_{\mathrm{pred}}(R)-\phi(R)^\top\hat a$ is the aligned residual that enters the energy channel of $s$. Because the correction is linear in composition alone, it removes the convention exactly and cannot absorb any geometry-dependent error; it can, however, absorb a composition-linear part of systematic model or reference-level bias, so it does not equate different electronic-structure methods. We fit $\hat a$ by iteratively trimmed least squares rather than plain least squares. A handful of MLIP blow-ups, reaching $|\Delta E|\sim10^{8}$\,eV on QCML, would otherwise dominate the normal equations, and the offset would not be recovered. Writing $\operatorname{TrimLS}_{I_{\mathrm{fit}}}(y)$ for this operator with $y_i=y(R_i)$ and $\phi_i=\phi(R_i)$: initialize $I_0=I_{\mathrm{fit}}$, the fitting configuration rows, and for $t=0,\ldots,5$ compute
\begin{equation*}
\begin{aligned}
 a^{(t)}&\in\arg\min_a\sum_{i\in I_t}(y_i-\phi_i^\top a)^2,
 \qquad e_i^{(t)}=y_i-\phi_i^\top a^{(t)},\\
 \tau_t&=\max\!\left\{5\,Q_{0.98}\!\left(\{|e_i^{(t)}|:i\in I_t\}\right),10^{-3}\,\mathrm{eV}\right\},
 \qquad J_t=\{i\in I_{\mathrm{fit}}:|e_i^{(t)}|<\tau_t\}.
\end{aligned}
\end{equation*}
Return the current coefficients if $|J_t|\le K+1$ or $t=5$; otherwise set $I_{t+1}=J_t$ and refit. The threshold quantile uses the current retained rows, but $J_t$ is evaluated over all original fitting rows, so a trimmed row can re-enter. At most six fits are performed.

\paragraph{Target residual normalization.} Because the three residual targets in Eq.~\eqref{eq:three_targets} can scale differently with atom count $N$,
we first remove their systematic size dependence. Let $r_t(R)$ denote the raw residual for target $t$, with $r_E(R)=|\Delta E(R)|$ the aligned energy residual.
For each target separately, the size exponent $\hat\beta_t$ is the least-squares slope, with intercept, of $\log r_t(R_i)$ on $\log N(R_i)$
over the fitting rows with $r_t(R_i)>0$. Writing $v_{it}=r_t(R_i)/N(R_i)^{\hat\beta_t}$, the scale is the root mean square over the rows at or below the $0.99$ quantile,
\begin{equation}\label{eq:scale}
q_t=Q_{0.99}\bigl(\{v_{it}:i\in I_{\mathrm{fit}}\}\bigr),
\quad
K_t=\{i\in I_{\mathrm{fit}}:v_{it}\le q_t\},
\quad
\hat c_t=
\left[\frac{1}{|K_t|}\sum_{i\in K_t}v_{it}^{2}\right]^{1/2},
\end{equation}
where $Q_{0.99}$ is the linearly interpolated empirical quantile with an inclusive boundary. Trimming estimates $\hat c_t$ only: it removes no configuration from any set and does not clip $u_t$. The true and predicted
residuals are then normalized using the same frozen quantities,
\begin{equation}
u_t(R)=
\frac{r_t(R)}
{\hat c_t N(R)^{\hat\beta_t}},
\qquad
\hat u_t(R)=
\frac{\hat r_t(R)}
{\hat c_t N(R)^{\hat\beta_t}},
\end{equation}
which removes systematic size dependence and places the three targets on
comparable numerical scales. The fitted $\hat\beta_t$ and $\hat c_t$ are then frozen throughout calibration and inference. The predictor outputs
$\hat u_t$ directly, so a predicted residual in physical units is recovered by
$\hat r_t(R)=\hat c_t N(R)^{\hat\beta_t}\hat u_t(R)$.

\paragraph{Committee-based scores.} We provide more details on how we calculate $\hat s$ for the committee baselines. The committee ranks configurations by disagreement among the three MLIP models used in Section~\ref{sec:experiments}. Let $m=1,\dots,M$ denote each model ($M=3$), and let $\hat e_m(R)$ and $\hat{\mathbf f}_{m,j}(R)\in\R^3$ be model $m$'s predicted total energy and force on configuration $R$. We use the same frozen public checkpoints for these models as in other experiments in the paper; the committee contains MACE-OFF23(M) as a member on the two MACE-OFF23 cells, where it is excluded as a target.

For each model, raw total energies differ in scale by composition-linear offsets---up to $1.9\times10^{5}$\,eV per iodine---that would dominate any genuine disagreement. We therefore place every member on a common scale before measuring the spread of predictions. Let $\phi(R)\in\R^{K+1}$ be the element-count vector of $R$, take member~$1$ as the reference, and fit
  \begin{equation}\label{eq:align}
    \hat b_m=\operatorname{TrimLS}_{I_{\mathrm{fit}}}(\hat e_1-\hat e_m),
    \qquad
    \tilde e_m(R)=\hat e_m(R)+\phi(R)^\top \hat b_m,
  \end{equation}
with $\tilde e_1=\hat e_1$, using the same robust trimming as the energy alignment; only model outputs and element counts enter Eq.~\eqref{eq:align}, so the aligned score remains label-free.
We then calculate the spread of $\tilde e_m(R)$ and $\hat{\mathbf f}_{m,j}$. The energy spread is
  \begin{equation}
    d_E(R)=\Bigl(\tfrac1M\sum\nolimits_m\bigl(\tilde e_m(R)-\bar e(R)\bigr)^2\Bigr)^{1/2},
    \qquad \bar e=\tfrac1M\sum\nolimits_m\tilde e_m,
  \end{equation}
and we summarize the atoms two ways to match the two force channels of $s$:
\begin{equation}
    d_{\mathrm{avg}}(R)=\frac1N\sum_{j=1}^{N}\sigma_j(R),
    \qquad
    d_{\max}(R)=\max_{j\in [N]}\ \sigma_j(R),
  \end{equation}
where
  \begin{equation}
    \sigma_j(R)=\Bigl(\tfrac1M\sum\nolimits_m
    \bigl\lVert \hat{\mathbf f}_{m,j}(R)-\bar{\mathbf f}_{j}\bigr\rVert_2^2\Bigr)^{1/2},
    \qquad \bar{\mathbf f}_{j}=\tfrac1M\sum\nolimits_m \hat{\mathbf f}_{m,j}.
  \end{equation}
The three channels carry incomparable units, so the \emph{raw committee} replaces each channel by its rank among the fitting configurations. With $n_{\mathrm{fit}}$ fitting configurations,
\begin{equation}\label{eq:frozen-committee}
\widehat F_{c,\mathrm{fit}}(x)=\frac{1}{n_{\mathrm{fit}}}
\sum_{R'\in\mathcal R_{\mathrm{fit}}}\mathbf 1\{d_c(R')\le x\},
\qquad
\hat s_{\mathrm{comm}}(R)=\sum_c w_c\widehat F_{c,\mathrm{fit}}(d_c(R)),
\end{equation}
with the same weights $w$ that define $s$ in Section~\ref{sec:error_predictor}; the inclusive inequality fixes the treatment of ties. The \emph{supervised committee} uses the same committee features but learns how to combine them: it replaces the fixed rank combination in Eq.~\eqref{eq:frozen-committee} with one residual-regression network, with the same hidden widths and output channels as an individual structural scorer, trained with seed $0$ on $\Dfit$ from the frozen channels $d_E,d_{\mathrm{avg}},d_{\max}$ with the $L_1$ objective. It thus tests whether a learned score on committee information can match the structural predictor. Both committees require three pretrained-MLIP evaluations per configuration.

We also experiment with two more baselines. \emph{Random} passes a frozen randomized ranking through CRC and serves as a lower reference. \emph{Oracle} passes the true score $\hat s=s$ through the same protocol and is retrospective: computing it requires exactly the labels the procedure is meant to avoid, so it is a cost reference rather than a deployable selector.

\paragraph{Electronic-structure references and force conventions.}\label{app:provenance}
The residuals of Eq.~\eqref{eq:residuals} compare each MLIP with the dataset's evaluation
reference. At least one component of that reference differs from the MLIP's training
reference in every combination (Table~\ref{tab:provenance}). Residuals therefore include
systematic differences between electronic-structure levels as well as MLIP approximation
error; the composition-linear alignment removes an offset but does not make functionals, basis sets or dispersion treatments equivalent, and force
residuals are not aligned. MACE-OFF23(M) is excluded as a target on the MACE-OFF23 population because of shared training-source provenance ($25$ exact configurations are recorded as shared with the associated training/validation source); complete training membership is not established for every model, and no blanket unseen or out-of-distribution claim is made.
All comparisons use physical forces $\mathbf F=-\nabla_R E$ in eV/\AA\ and energies in eV. COMP6's upstream array named \texttt{forces} stores the energy gradient, $+\nabla_R E$; ingestion negates that raw vector exactly and converts Hartree/\AA\ to eV/\AA\ with factor $27.211386245988$. MACE-OFF23 and QCML references already have physical-force semantics and receive no sign change. Absolute energies are never pooled across different electronic-structure references.

\begin{table}[htbp]
\centering\small
\setlength{\tabcolsep}{4pt}
\renewcommand{\arraystretch}{1.08}
\resizebox{\linewidth}{!}{%
\begin{tabular}{llll}
\toprule
dataset & evaluation reference & target MLIP & MLIP training reference \\
\midrule
COMP6 & $\omega$B97X/6-31G(d), Gaussian 09 & MACE-OFF23(M) & $\omega$B97M-D3(BJ)/def2-TZVPPD \\
 & & AIMNet2 & $\omega$B97M-D3/def2-TZVPP \\
 & & AceFF-2 & $\omega$B97M-V/def2-TZVPPD, VV10 \\
\addlinespace
QCML & PBE0, FHI-aims & MACE-OFF23(M) & $\omega$B97M-D3(BJ)/def2-TZVPPD \\
 & & AIMNet2 & $\omega$B97M-D3/def2-TZVPP \\
 & & AceFF-2 & $\omega$B97M-V/def2-TZVPPD, VV10 \\
\addlinespace
MACE-OFF23 & $\omega$B97M-D3(BJ)/def2-TZVPPD, PSI4 & AIMNet2 & $\omega$B97M-D3/def2-TZVPP \\
 & & AceFF-2 & $\omega$B97M-V/def2-TZVPPD, VV10 \\
\bottomrule
\end{tabular}%
}
\caption{\textbf{Electronic-structure references.} Each dataset is paired with the reference used to train each target MLIP; at least one component differs in every combination \citep{smith2018less,kovacs2025maceoff,anstine2025aimnet2,aceff2,ganscha2025qcml}.}
\label{tab:provenance}
\end{table}

\section{The shared guarantee}\label{app:guarantee}

This section extends the top row of Fig.~\ref{fig:main} to all eight cells and asks what the finite-sample term of Definition~\ref{def:cutoff} costs. The risk ratio $r(\alpha)$ and the realized retained loss $\rho(\alpha)$ are defined in Section~\ref{sec:exp:setup} and Appendix~\ref{app:metrics}.

\begin{figure}[htbp]
\centering\includegraphics[width=\linewidth]{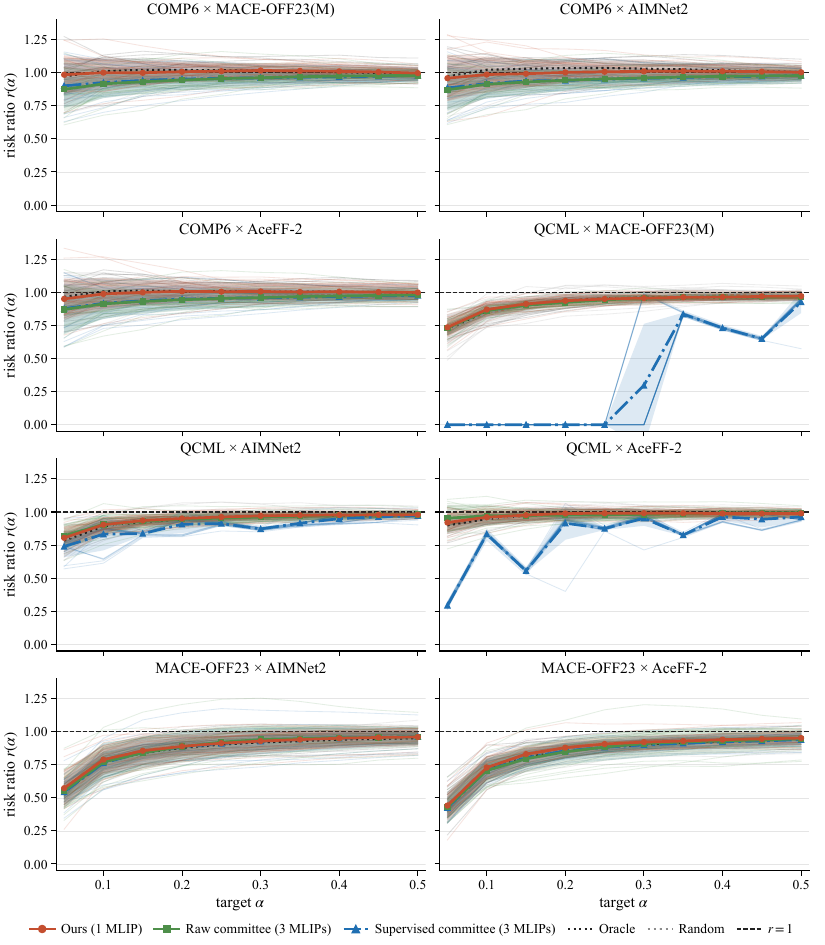}
\caption{\textbf{Every score meets the same risk guarantee in every cell.} Risk ratio $r(\alpha)$ (realized capped test loss over the budget $\epsilon_0=\alpha\bar\ell_{\rm fit}$) for all eight cells and five scores. Faint lines are the $20$ calibration/test reallocations; bold lines and bands are means $\pm1$ SD; the dashed line marks $r=1$. Single draws above $1$ are expected because the bound is on the expectation. On QCML $\times$ MACE-OFF23(M) the supervised committee routes the whole pool for $\alpha\le0.25$, so its $r$ is exactly $0$ there.}
\label{fig:app-E1}
\end{figure}

\begin{figure}[htbp]
\centering\includegraphics[width=\linewidth]{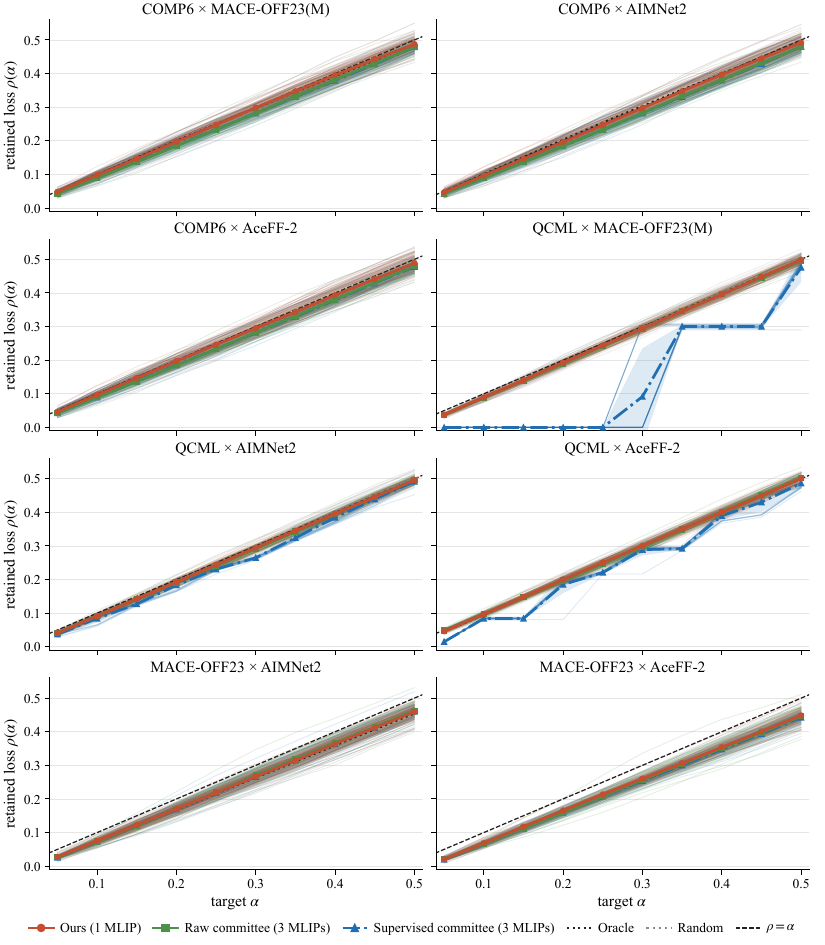}
\caption{\textbf{The realized retained loss tracks the target for every score.} Retained fraction $\rho(\alpha)$, the capped loss left uncorrected over the pool's own total, for all eight cells; the dashed line is $\rho=\alpha$, a visual reference rather than a guarantee. Faint lines: the $20$ reallocations; bold: mean $\pm1$ SD. Points below the diagonal (MACE-OFF23 cells; the tied supervised committee on QCML) mean that more DFT was spent than the target required.}
\label{fig:app-E2}
\end{figure}

Across all cells, scores, targets, and reallocations, the largest mean risk ratio over all reallocations is $\RiskMeanMax$ and the largest single draw is $\RiskMax$ (Fig.~\ref{fig:app-E1}); the guarantee bounds an expectation, so individual draws and finite means may exceed $1$. At small $\alpha$ the ratio falls well below $1$ for every score. The cutoff keeps the calibration loss under $(n{+}1)\epsilon_0-B$, and $\epsilon_0=\alpha\,\bar\ell_{\rm fit}$ shrinks with $\alpha$ while the identity bound $B$ does not, so the fixed term $B/(n{+}1)$ consumes most of a small budget. The retained fraction $\rho(\alpha)$ tracks $\alpha$ for every score (Fig.~\ref{fig:app-E2}); points below the diagonal mean that the rule spent more DFT than the target required.

\begin{figure}[htbp]
\centering\includegraphics[width=\linewidth]{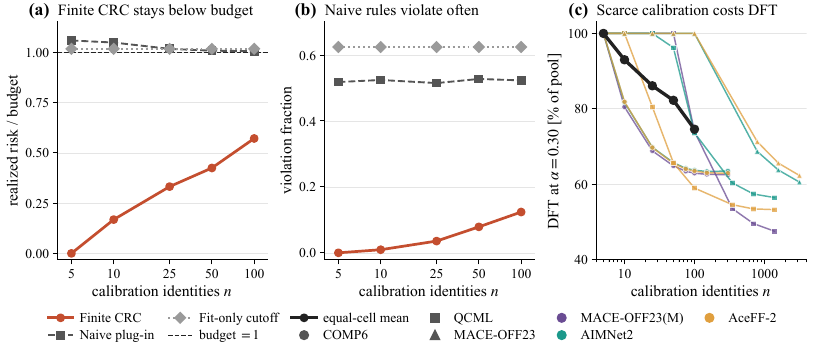}
\caption{\textbf{The finite correction keeps the budget; its price is DFT when calibration is scarce.} (a) Realized risk/budget and (b) fraction of calibration draws exceeding the budget at $\alpha=0.30$, for finite CRC, a plug-in cutoff without $B/(n{+}1)$, and a cutoff fixed on $\Dfit$; equal-cell means over eight cells ($200$ draws for $n\le100$). (c) Deployment DFT at $\alpha=0.30$: faint lines are cells on their own calibration grids, the bold line the equal-cell mean over the five shared $n$. The ranking score is fixed throughout; log $x$ axes.}
\label{fig:app-E3}
\end{figure}

Fig.~\ref{fig:app-E3} holds the ranking score fixed and varies the calibration rule and the number $n$ of calibration identities, at $\alpha=0.30$. A plug-in cutoff that omits $B/(n{+}1)$ has mean risk/budget $1.01$--$1.06$ for $n=5$--$100$ and exceeds the budget in about $52\%$ of calibration draws; a cutoff fixed on $\Dfit$ alone exceeds it in $62\%$. Finite CRC stays below budget on average ($0.17$ and $0.9\%$ violations at $n=10$) but routes the whole pool at $n=5$ and $74.6\%$ at $n=100$, against $\OursDFT\%$ at the full calibration size. Violation frequencies describe the observed draws; CRC makes no high-probability claim.

\FloatBarrier

\section{DFT cost in every cell}\label{app:cost}

This section extends the middle and bottom row of Fig.~\ref{fig:main} and Fig.~\ref{fig:gap}(b) to every cell. Table~\ref{tab:main} gives the equal-cell means, Table~\ref{tab:score-summary} the per-cell $G$, and Table~\ref{tab:matched-resources} the matched-resource comparisons.

\begin{figure}[htbp]
\centering\includegraphics[width=\linewidth]{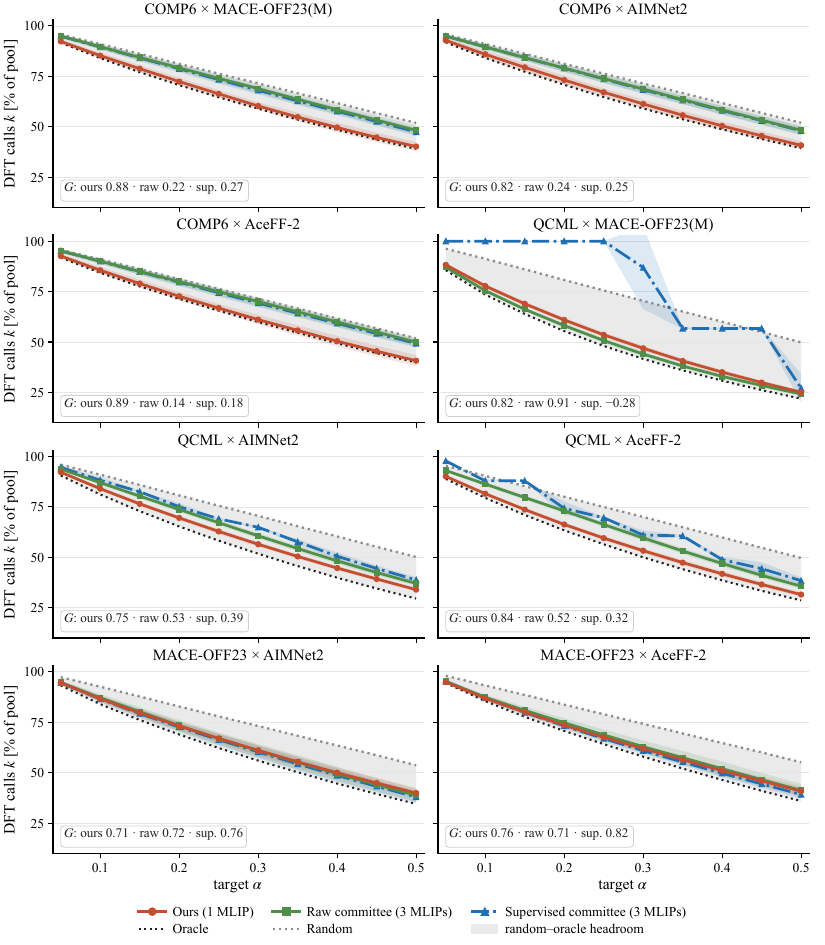}
\caption{\textbf{Within the shared guarantee, ours needs the fewest DFT calls of the deployable scores in most cells.} DFT calls $k(\alpha)$ for all eight cells; the gray band is the random-to-oracle headroom and each panel prints $G$ for ours and the two committees. Means over $20$ reallocations with $\pm1$ SD bands for ours and the committees. The raw committee wins QCML $\times$ MACE-OFF23(M) and the committees match or beat ours on MACE-OFF23. The oracle is retrospective, not deployable.}
\label{fig:app-F1}
\end{figure}

\begin{figure}[htbp]
\centering\includegraphics[width=\linewidth]{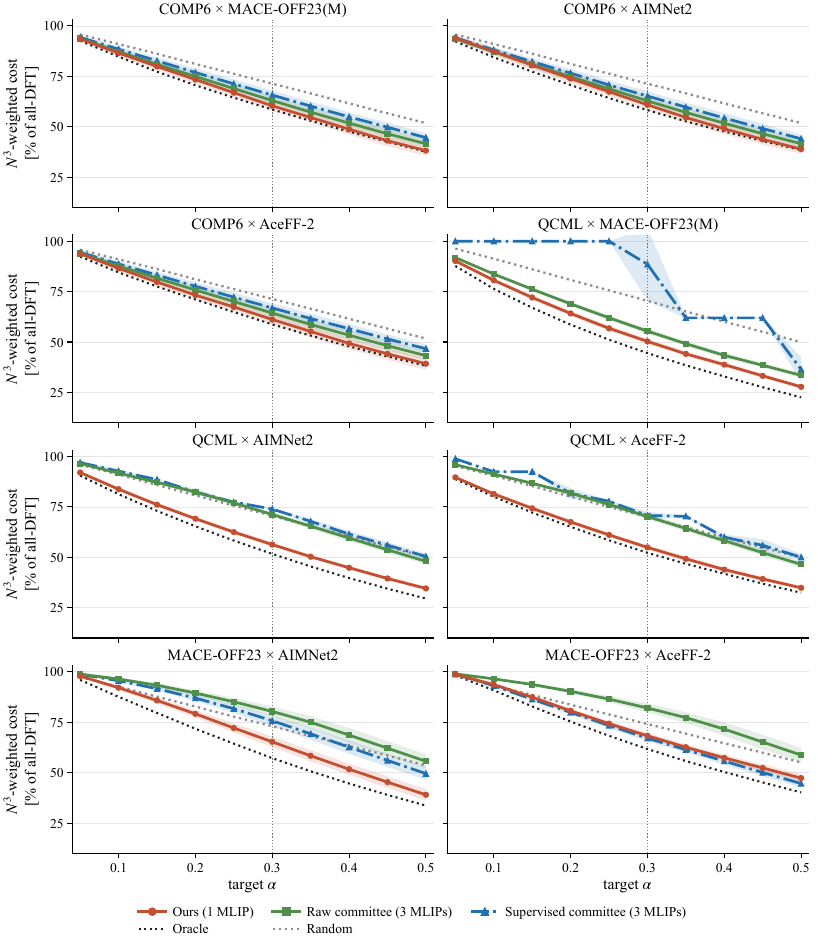}
\caption{\textbf{Weighting each DFT call by $N^3$ preserves, and on QCML widens, the advantage of ours.} $N^3$-weighted DFT cost, the summed cubed atom counts of routed configurations over the same sum for the pool, for all eight cells; the dotted line marks $\alpha=0.30$. Means over $20$ reallocations with $\pm1$ SD bands for ours and the committees. The proxy captures size scaling, not measured wall time, and the comparison fixes $\alpha$, not the realized loss.}
\label{fig:app-F2}
\end{figure}

\begin{figure}[htbp]
\centering\includegraphics[width=\linewidth]{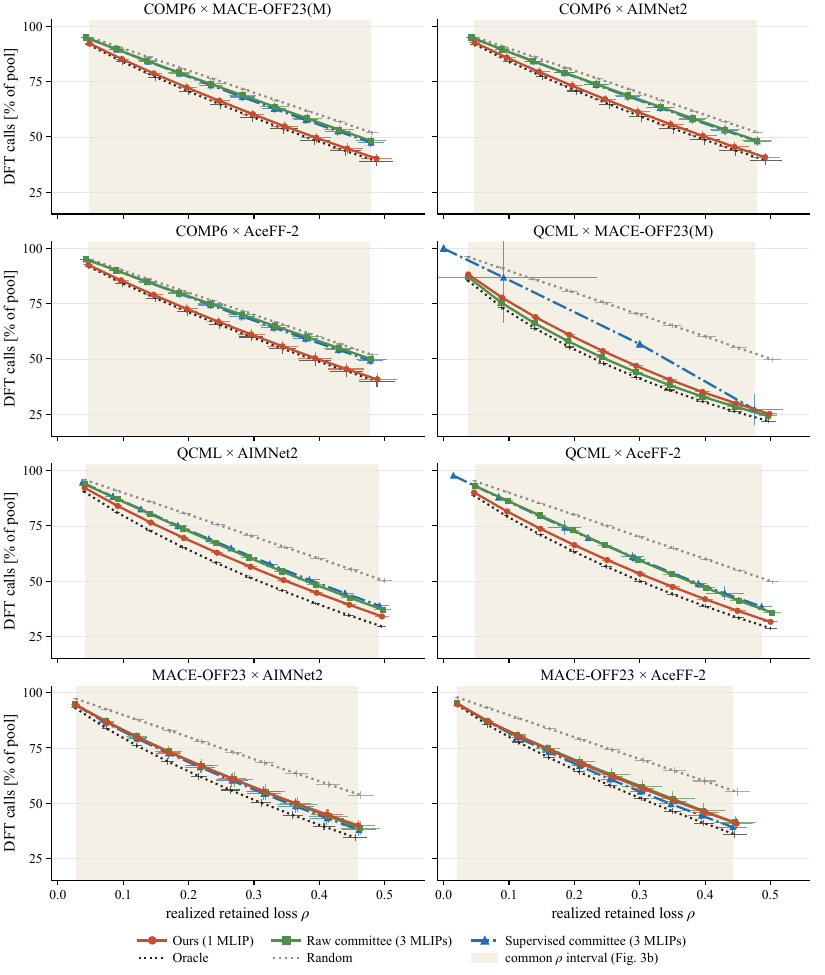}
\caption{\textbf{At equal realized loss, ours still calls DFT least in most cells.} DFT calls against realized retained loss $\rho$; each point is one target $\alpha$ at the $20$-reallocation mean, with $\pm1$ SD whiskers in both coordinates. The shaded interval is the common attained $\rho$ range of ours and the two committees, over which Fig.~\ref{fig:gap}(b) averages the DFT difference. On QCML $\times$ MACE-OFF23(M), the supervised committee's first five $\alpha$ values collapse to one all-DFT point at $\rho=0$.}
\label{fig:app-F3}
\end{figure}

At $\alpha=0.30$, ours routes less DFT than both committees in five of eight cells (Fig.~\ref{fig:app-F1}); the raw committee is closer to the oracle on QCML $\times$ MACE-OFF23(M), and the committees match or beat ours on the two MACE-OFF23 cells. On QCML $\times$ MACE-OFF23(M) the supervised committee routes the entire pool for $\alpha\le0.25$, and in $\SupWholePoolRepeats$ of $20$ reallocations at $\alpha=0.30$. Note the rule permits this conservative fallback, and it is why that score has $G<0$. Weighting each call by $N^3$ (Fig.~\ref{fig:app-F2}), ours is cheaper than the raw committee and random in all eight cells for $\alpha\ge0.15$ and cheaper than the supervised committee in seven, the exception being MACE-OFF23 $\times$ AceFF-2. At equal realized loss (Fig.~\ref{fig:app-F3}), ours saves DFT against each committee in six of eight cells.

\begin{table}[htbp]
\centering\small
\input{tables/T_main}
\caption{\textbf{Main results, equal-weight means over the eight cells.} $G$ is the normalized gain Eq.~\eqref{eq:gain}; DFT and $\rho$ are the deployment DFT fraction and realized retained loss at $\alpha=0.30$. The last three columns compare each committee with ours over the common attained support of every cell: DFT saved by ours at matched $\rho$ (Appendix~\ref{app:matched-resource}), and the relative-energy MAE and selection-success change of ours at matched deployment DFT (Appendix~\ref{app:physical-grid}).}
\label{tab:main}
\end{table}

\begin{table}[htbp]
\centering\small\input{tables/T5_score_summary}
\caption{\textbf{Normalized gain $G$ per cell.} Each $G$ uses the $20$-reallocation mean DFT-call curve on the $\alpha$ grid. Among ours and the two committees, bold marks the best and underlining the second best; the last three columns are label-free controls (Appendix~\ref{app:baselines-strength}), which never exceed ours. The final row is the equal-cell mean.}\label{tab:score-summary}
\end{table}

\subsection{Matched-resource comparisons}\label{app:matched-resource}
Equal $\alpha$ does not imply equal realized loss: at $\alpha=0.30$ the supervised committee retains $\rho=\SupRho$ against $\OursRho$ for ours (Table~\ref{tab:main}), so part of its extra DFT buys a lower retained loss. For each cell we therefore form the $20$-reallocation mean curves at the ten $\alpha$ values. The common $\rho$ interval $I_\rho$ is the intersection of the ranges attained by ours and the two committees; random, oracle, and the all-MLIP and all-DFT endpoints do not extend it. Within $I_\rho$, piecewise-linear interpolation defines each curve, and for baseline $b$
\begin{equation}
\overline{\Delta\mathrm{DFT}}_b=
\frac{1}{|I_\rho|}\int_{I_\rho}
[\mathrm{DFT}_b(\rho)-\mathrm{DFT}_{\rm ours}(\rho)]\,d\rho ,
\end{equation}
integrated exactly over the union of curve knots, so positive values favor ours. Matched deployment DFT uses the analogous common DFT interval, with differences $\mathrm{MAE}_{\rm ours}-\mathrm{MAE}_b$ and $\mathrm{Success}_{\rm ours}-\mathrm{Success}_b$ (Appendix~\ref{app:physical-grid}). Cell values are averaged with equal weight. No extrapolation or endpoint extension is used, and these are descriptive comparisons over common support, not new CRC guarantees.

\begin{table}[htbp]
\centering\small\input{tables/T_matched_resources}
\caption{\textbf{Matched resources over common attained support.} (a) Comparisons with the raw and supervised committees; arrows mark the direction favoring ours. DFT saving is baseline minus ours; MAE and success changes are ours minus baseline. The bold mean weights the eight cells equally; pp denotes percentage points. (b) Common intervals used for the comparisons.}
\label{tab:matched-resources}
\end{table}
\FloatBarrier

\section{Why the gap opens and closes}\label{app:mechanism}

This section tests the reading of Fig.~\ref{fig:gap}(a): the structural score wins where the MLIPs share error with one another, and disagreement wins where failures are rare, catastrophic, and model-specific.

\begin{figure}[htbp]
\centering\includegraphics[width=\linewidth]{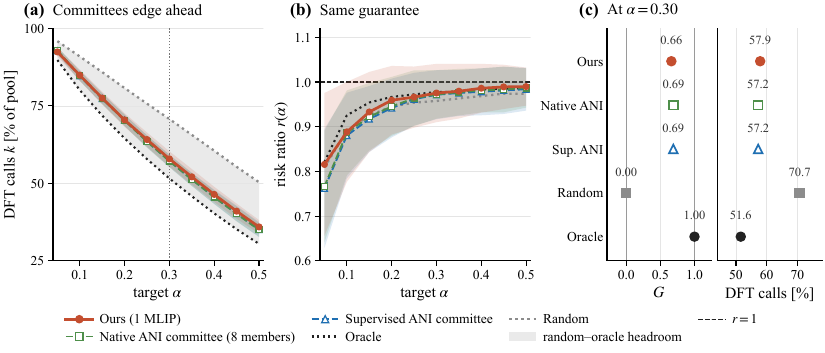}
\caption{\textbf{With a matched reference, committees edge ahead, and the structural score stays far above random.} ANI-2x $\times$ COMP6: (a) DFT calls with the random-to-oracle headroom shaded; (b) risk ratio $r(\alpha)$; (c) $G$ and DFT calls at $\alpha=0.30$ for ours, the native ANI committee (members $0$--$7$; target is their mean), and the supervised ANI committee. Means over $20$ reallocations with $\pm1$ SD bands for the three selectors. ANI-2x training membership for these configurations is unverified, so this is not an unseen-data test.}
\label{fig:app-G1}
\end{figure}

\paragraph{Matched reference.} ANI-2x is a different MLIP whose eight released members were trained at the nominal COMP6 reference, $\omega$B97X/6-31G(d). We use the mean of members $0$--$7$ as the target, the native committee is their disagreement (population standard deviation), and the supervised ANI committee applies the supervised-committee construction to the same ensemble. If disagreement misses shared error, it should recover here. It does, slightly (Fig.~\ref{fig:app-G1}): at $\alpha=0.30$ ours has $G=0.66$ and routes $57.9\%$, against $0.69$ and $57.2\%$ for both ANI committees and $70.7\%$ for random, and at matched $\rho$ ours needs $0.71$ and $0.75$ pp more DFT. Training membership of these configurations could not be established, so this is not an unseen-data test.

\begin{figure}[htbp]
\centering\includegraphics[width=\linewidth]{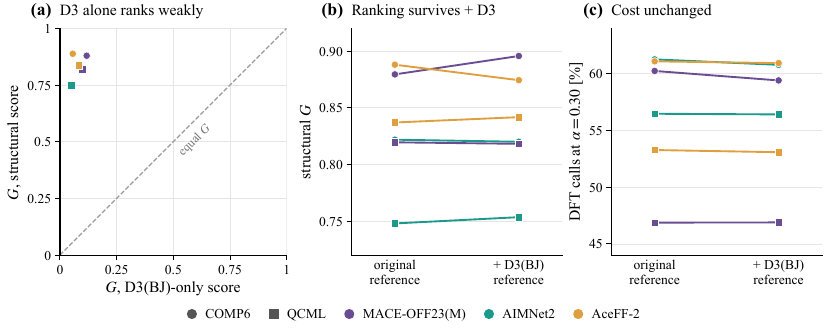}
\caption{\textbf{An isolated D3(BJ) term does not explain the structural ranking.} (a) Structural $G$ against the $G$ of a D3(BJ)-only score (dashed: equal $G$). (b) Structural $G$ and (c) DFT calls at $\alpha=0.30$ with the original reference and after adding D3(BJ) to the reference, joined per cell. Six COMP6 and QCML cells; cell statistics over $20$ reallocations. Adjusting the reference changes the loss and refits the predictor, so before and after $G$ use their own random and oracle anchors. A diagnostic, not proof that reference mismatch is absent.}
\label{fig:app-G2}
\end{figure}

\paragraph{Isolated dispersion control.} D3(BJ) is an empirical correction for long-range dispersion that the evaluation and training references treat differently. A score built from the D3(BJ) term alone reaches $G=0.05$--$0.12$ in the six COMP6 and QCML cells, while adding D3(BJ) to the reference and refitting leaves structural $G$ at $0.75$--$0.90$, with each cell changing by at most $0.02$ and DFT at $\alpha=0.30$ by under $1$ pp (Fig.~\ref{fig:app-G2}). This is a diagnostic control, not proof that all electronic-structure mismatch is absent: functional form, basis set, and other effects remain.

\begin{figure}[htbp]
\centering\includegraphics[width=\linewidth]{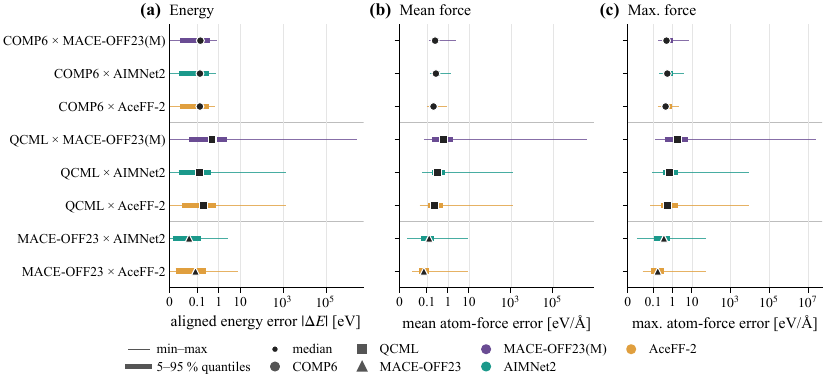}
\caption{\textbf{Residual scales differ by orders of magnitude across cells, with catastrophic tails only on QCML.} Aligned energy error, mean atom-force error, and maximum atom-force error of the target MLIP on the complete resample-$0$ test pools. Thin lines span minimum to maximum, thick lines the $5$th--$95$th percentiles, markers the medians. Asinh axes, linear below $0.01$, so zero and the largest QCML $\times$ MACE-OFF23(M) force error ($2.5\times10^{7}$ eV/\AA) are both visible; nothing is omitted.}
\label{fig:app-G5}
\end{figure}

\paragraph{Residual concentration.} The Gini coefficient of the pool's residual mass is $\le0.18$ on COMP6, where error is diffuse, and $1.00$ on QCML $\times$ MACE-OFF23(M), where a handful of configurations carry essentially all of it (Table~\ref{tab:headroom-app}, Fig.~\ref{fig:app-G5}). The latter is the one cell where the raw committee beats ours.

\paragraph{Shared and target-specific components.} For each configuration, the \emph{consensus} residual is the mean of the three MLIPs' signed aligned energy residuals, or of their per-atom force-residual vectors, and the \emph{target-specific} residual is the target's residual minus that consensus, with force norms taken after vector subtraction. On the resample-$0$ test pools, the mean Spearman correlation of ours over the eight cells is $0.47$/$0.54$/$0.45$ with the consensus energy, mean-force, and maximum-force magnitudes, against $0.28$/$0.45$/$0.36$ with the target-specific ones. The raw committee correlates more with the target-specific force components ($0.89$/$0.86$) than with the consensus ones ($0.65$/$0.62$). The target-specific part equals the deviation of the target's prediction from the ensemble mean, so the committee's association with it is partly built in; these are descriptive associations, not a causal decomposition.

\begin{figure}[htbp]
\centering\includegraphics[width=\linewidth]{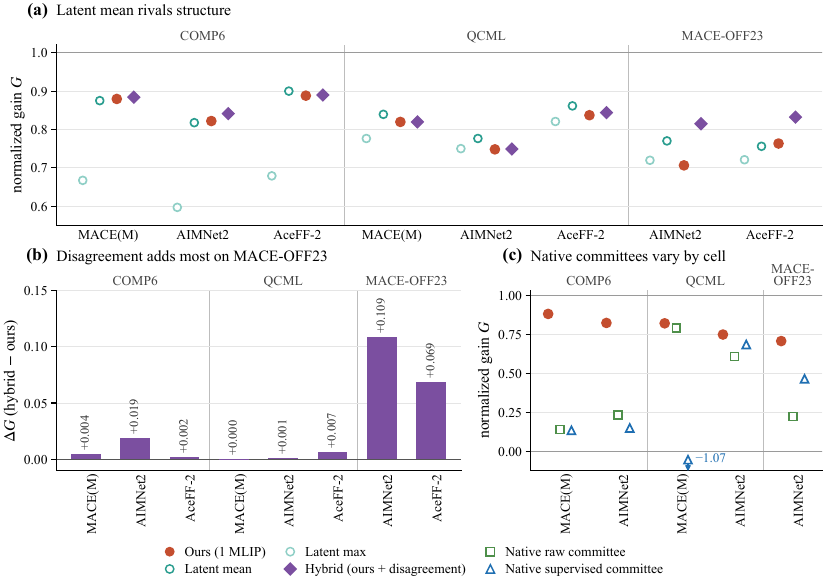}
\caption{\textbf{Target-model latents rival but do not dominate the structural score, and disagreement adds most on MACE-OFF23.} (a) Per-cell $G$ for ours, latent mean, latent max, and the hybrid ($y$ axis zoomed to $0.55$--$1$). (b) Hybrid minus ours. (c) Native raw and supervised committees against ours in the five cells where they are defined; the native supervised value $-1.07$ on QCML $\times$ MACE-OFF23(M) is drawn clipped. Cell statistics over $20$ reallocations. The small QCML hybrid gains are not claimed as improvements.}
\label{fig:app-G3}
\end{figure}

\begin{figure}[htbp]
\centering\includegraphics[width=\linewidth]{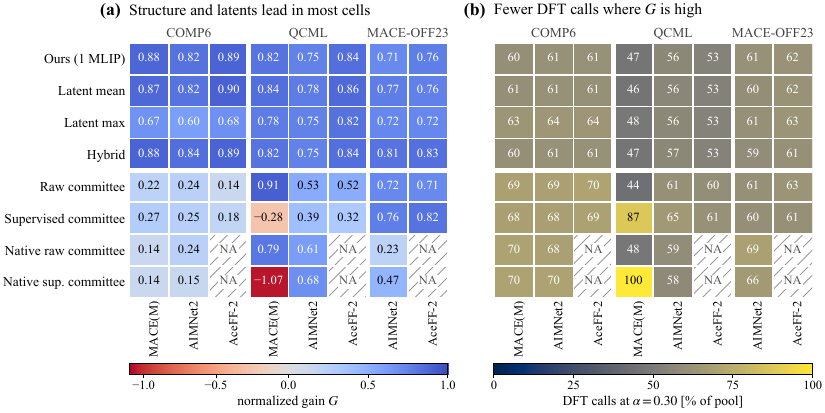}
\caption{\textbf{Every score source in every cell.} (a) $G$ (diverging map centered at $0$) and (b) DFT calls at $\alpha=0.30$ (fixed $0$--$100$ scale) for ours, latent mean and max, the hybrid, the raw and supervised committees, and the native raw and supervised committees. Cell statistics over $20$ reallocations. Hatched NA entries are undefined native committees, not zeros.}
\label{fig:app-G4}
\end{figure}

\paragraph{Model internals and hybrids.} The \emph{latent} scores feed the target MLIP's per-atom invariant features ($256$ dimensions), mean- or max-pooled over atoms, to the same predictor and CRC protocol as ours. The \emph{hybrid} appends the three disagreement channels of the raw committee to our 350-dimensional descriptor. \emph{Native} committees use members of the target's own family: AIMNet2 members $0$--$3$ with member $0$ as the frozen target, and MACE-OFF23 S/M/L with M as the target; they are undefined for AceFF-2 targets and for MACE-OFF23(M) on the MACE-OFF23 population. Latent mean pooling (mean $G=0.82$) exceeds ours in five of eight cells, while max pooling ($0.72$) drops to $0.60$--$0.68$ on COMP6 (Fig.~\ref{fig:app-G3}). The hybrid ($0.83$) is at or above ours in every cell, by $+0.11$ and $+0.07$ on MACE-OFF23 and by at most $0.02$ elsewhere; the QCML gains are too small to call improvements. Native committees reach only $0.14$--$0.24$ on COMP6, and the native supervised committee has $G=-1.07$ on QCML $\times$ MACE-OFF23(M). Fig.~\ref{fig:app-G4} gives every source in every cell.

\begin{table}[htbp]
\centering\small\resizebox{\linewidth}{!}{\input{tables/T4_headroom}}
\caption{\textbf{Residual concentration and frozen loss bounds.} Resample-$0$ test pool. Gini uses untruncated configuration scores $s$; ``Top 10\%'' is the untruncated score mass in the highest-mass $10\%$ of identities. $m_{\max}$ is the maximum fitting-identity size, $B_{\mathrm{cfg}}$ the configuration truncation level, and $B$ the identity bound of Eq.~\eqref{eq:loss}. The last two columns are in meV.}
\label{tab:headroom-app}
\end{table}

\FloatBarrier

\section{Are the committees and controls strong enough?}\label{app:baselines-strength}

\begin{figure}[htbp]
\centering\includegraphics[width=\linewidth]{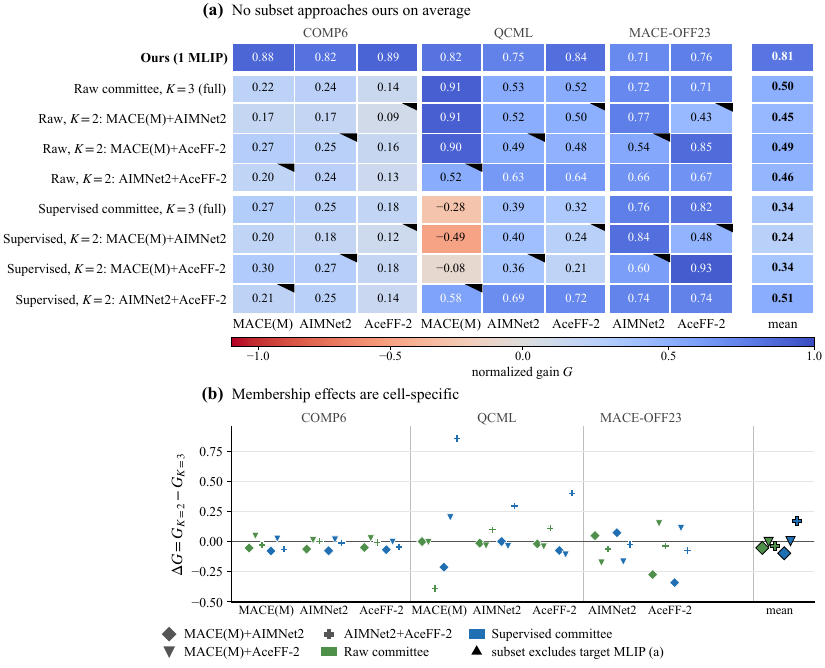}
\caption{\textbf{No two-member committee approaches the structural score on average; membership effects are cell-specific.} (a) $G$ for ours, the full raw and supervised committees ($K{=}3$), and each two-member subset ($K{=}2$), per cell and as the equal-cell mean. A black corner triangle marks subsets that exclude the target MLIP; these still need three distinct models, so they are not cheaper than $K{=}3$. (b) $\Delta G=G_{K=2}-G_{K=3}$ per cell and subset, raw (green) and supervised (blue), with equal-cell means at right.}
\label{fig:app-H1}
\end{figure}

This section asks whether the gap in Fig.~\ref{fig:gap}(a,\,c) reflects weak baselines. Two-member committees keep the target and one other model, or drop the target; the latter still needs three distinct MLIPs, since the target must be evaluated. No two-member subset approaches ours on average (Fig.~\ref{fig:app-H1}): mean $G$ is $\RawKTwoMin$--$\RawKTwoMax$ for raw and $\SupKTwoMin$--$\SupKTwoMax$ for supervised subsets, against $\OursGainThree$ for ours. Membership effects are cell-specific; the supervised AIMNet2 + AceFF-2 subset gains $+0.17$ on average, driven by QCML $\times$ MACE-OFF23(M) ($+0.86$), where the full supervised committee is below random. Some subsets beat ours in individual QCML $\times$ MACE-OFF23(M) and MACE-OFF23 cells.

\begin{figure}[htbp]
\centering\includegraphics[width=\linewidth]{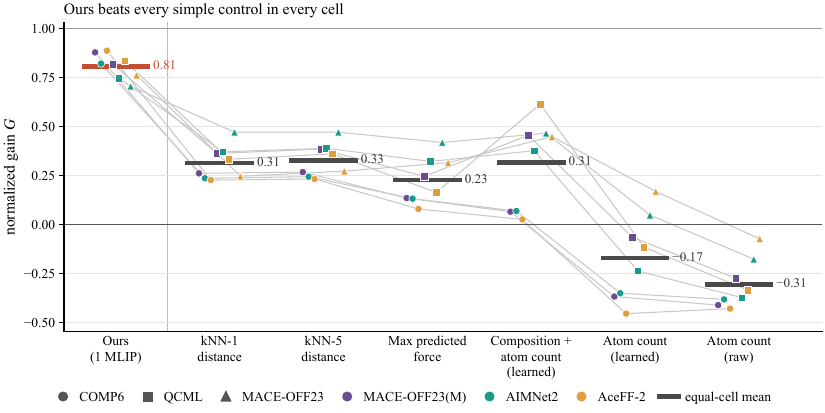}
\caption{\textbf{The structural score beats every simple model-agnostic control in every cell.} Per-cell $G$ for ours and six controls under the same loss and CRC rule: kNN-1 and kNN-5 descriptor-space distance, the target MLIP's maximum predicted force, learned composition + atom count, learned atom count, and raw atom count. Gray lines join the same cell (marker: dataset; color: MLIP); bars are equal-cell means. Negative $G$ (worse than random) is shown unclipped and does not imply that size is irrelevant.}
\label{fig:app-H2}
\end{figure}

The model-agnostic controls pass through the same loss and CRC rule (Fig.~\ref{fig:app-H2}): \emph{kNN-1} and \emph{kNN-5}, the descriptor-space distance to the nearest fitting configuration or the mean distance to the nearest five ($G=\KnnOneGain$ and $\KnnFiveGain$); \emph{max predicted force}, the target MLIP's largest predicted atomic force magnitude ($\NativeForceGain$); a \emph{composition + atom count} score and a \emph{learned atom count} score, each trained with our predictor's targets and objective on element counts and atom count, or atom count alone ($0.31$ and $-0.17$); and \emph{raw atom count} ($-0.31$). Ours exceeds every control in every cell. Composition and size carry some signal but do not account for the structural gain.

\FloatBarrier

\section{What gives the structural score its ranking}\label{app:diagnostics}

\begin{figure}[htbp]
\centering\includegraphics[width=0.9\linewidth]{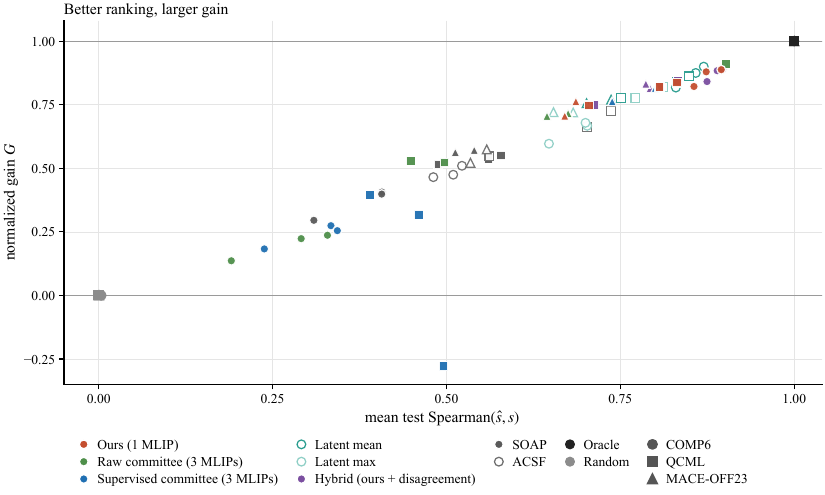}
\caption{\textbf{Scores that rank configurations better capture more of the random-to-oracle headroom.} $G$ against the mean test Spearman correlation between score and true error, per cell (marker: dataset) and score (color): the five main scores plus latent mean, latent max, hybrid, SOAP, and ACSF (hollow). Spearman values are means over $20$ test reallocations; $G$ uses the mean DFT-call curve. Oracle and random sit near $(1,1)$ and $(0,0)$.}
\label{fig:app-I1}
\end{figure}

\begin{figure}[htbp]
\centering\includegraphics[width=\linewidth]{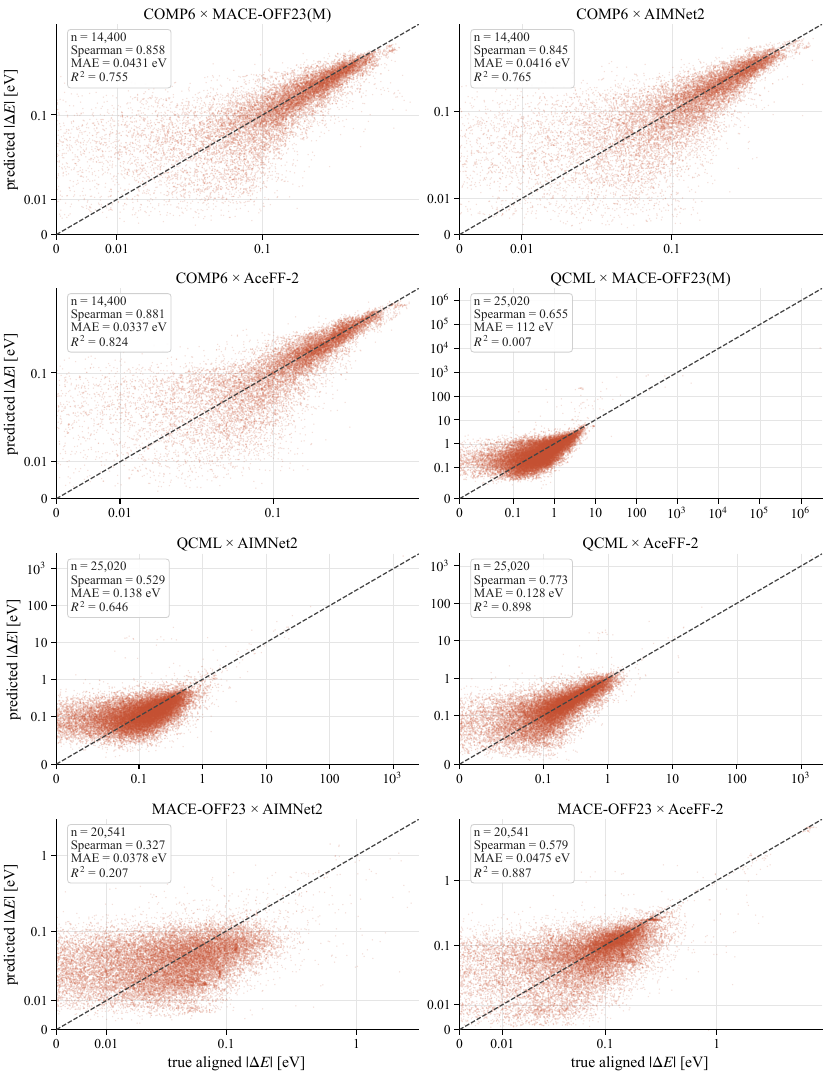}
\caption{\textbf{The predictor ranks energy residuals even where it misestimates their size.} True against predicted aligned $|\Delta E|$ (energy channel of the frozen predictor) for every resample-$0$ test configuration; asinh axes, linear below $0.01$ eV; dashed line $y=x$. Boxes give $n$, Spearman correlation, MAE, and $R^2$ on raw residuals with outliers kept. The QCML $\times$ MACE-OFF23(M) axis reaches about $10^{6}$ eV because of a few catastrophic residuals, which is also why its $R^2$ is near $0$.}
\label{fig:app-I2}
\end{figure}

This section explains Fig.~\ref{fig:gap}(c). CRC uses a score only through the order it induces, so ranking quality is what becomes DFT savings. Across all scores and cells, $G$ rises with the mean test Spearman correlation between $\hat s$ and $s$ (Fig.~\ref{fig:app-I1}). The predictor ranks residuals even where magnitude regression fails, as on QCML $\times$ MACE-OFF23(M), where energy $R^2$ is $0.007$ but the Spearman correlation is $0.66$ (Fig.~\ref{fig:app-I2}).

\begin{figure}[htbp]
\centering\includegraphics[width=\linewidth]{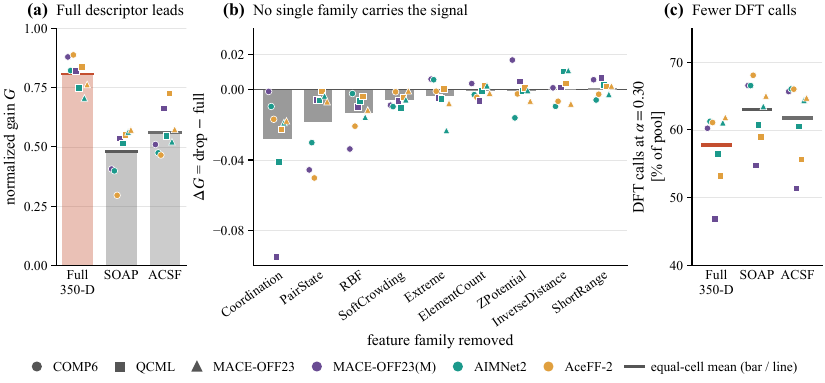}
\caption{\textbf{The full descriptor outperforms SOAP and ACSF, and no single feature family carries the signal.} (a) Equal-cell mean $G$ (bars) and per-cell values (dots) for the full 350-dimensional descriptor, SOAP, and ACSF. (b) $\Delta G$ (drop minus full) for each single-family drop, ordered by mean effect. (c) DFT calls at $\alpha=0.30$ ($y$ axis zoomed to $40$--$75\%$). All use the frozen fitting protocol and $20$ reallocations; small positive $\Delta G$ values are not significant improvements.}
\label{fig:app-I3}
\end{figure}

\begin{figure}[htbp]
\centering\includegraphics[width=\linewidth]{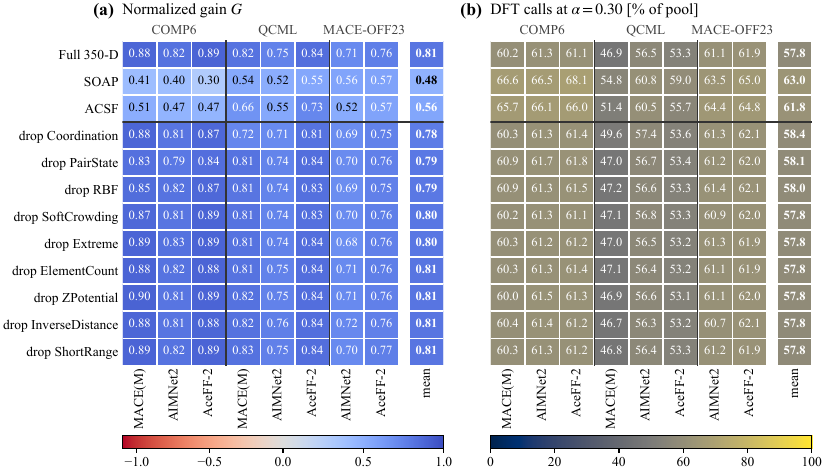}
\caption{\textbf{Descriptor effects are consistent across cells.} (a) $G$ (diverging scale centered at $0$) and (b) DFT calls at $\alpha=0.30$ (fixed $0$--$100$ scale) for every representation and single-family drop in every cell; the last column is the equal-cell mean. Rows below the rule are drops from the full descriptor, ordered by mean $\Delta G$.}
\label{fig:app-I4}
\end{figure}

\paragraph{Representation.} The full descriptor reaches mean $G=\OursGain$ against $0.56$ for ACSF and $0.48$ for SOAP, and routes $\OursDFT\%$ at $\alpha=0.30$ against $61.8\%$ and $63.0\%$ (Figs.~\ref{fig:app-I3} and~\ref{fig:app-I4}). This concerns the tested residual-ranking task, not descriptor quality in general. Removing one of the nine feature families changes mean $G$ by at most $0.03$ (Coordination, $-0.028$; PairState, $-0.019$; RBF, $-0.013$; the rest within $\pm0.006$), so the ranking information is spread across families; small positive changes are point estimates, not improvements.

\begin{figure}[htbp]
\centering\includegraphics[width=\linewidth]{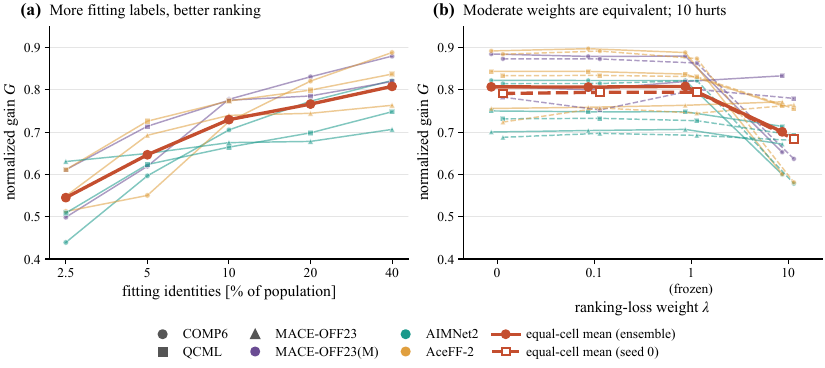}
\caption{\textbf{More fitting labels improve ranking; the ranking-loss weight matters only when large.} (a) $G$ against fitting-set size ($2.5$--$40\%$ of population identities, nested subsets of the original fitting pool), one thin line per cell and the bold equal-cell mean. (b) $G$ against the ranking-loss weight $\lambda\in\{0,0.1,1\text{ (frozen)},10\}$ for the three-seed ensemble (solid) and seed $0$ (dashed). Lines join cell values and are not uncertainty intervals.}
\label{fig:app-I5}
\end{figure}

\paragraph{Training choices.} More fitting labels improve ranking: mean $G$ rises from $0.54$ with $2.5\%$ of population identities to $0.81$ with $40\%$ (Fig.~\ref{fig:app-I5}a). Each fitting size refits the loss quantities as well as the predictor, so each size defines its own proxy loss. The pairwise ranking loss does not materially help at the frozen weight: $\lambda_{\rm rank}\in\{0,0.1,1\}$ give mean $G=0.807$, $0.806$, and $0.808$, and $\lambda_{\rm rank}=10$ gives $0.70$ (Fig.~\ref{fig:app-I5}b). The frozen $\lambda_{\rm rank}=1$ is retained because it was selected before test evaluation.

\paragraph{Fitting-only selection.}\label{app:score-validation}
Architecture, hyperparameters, the three seeds, $\lambda_{\rm rank}=1$, and arithmetic averaging were chosen on $\Dfit$ and its fixed internal validation rows by the equal-cell mean Spearman correlation with the capped loss $\ell$, then frozen before calibration and testing. The internal split is by configuration, so validation identities also appear in training rows; it supports selection, not an identity-level generalization test. The diagnostics above were computed after freezing and did not enter selection.

\FloatBarrier

\section{What routing does to chemical observables}\label{app:physical-grid}

This section extends the relative-energy results of Section~\ref{sec:exp:main}. These quantities are outside Proposition~\ref{thm:crc}.

\paragraph{Definitions.} Deployment DFT is the percentage of the test pool routed to DFT. Hybrid energies take the reference energy for routed configurations and the MLIP energy, with the frozen fitting-only alignment, otherwise; the all-MLIP endpoint uses raw predictions, since a composition-linear offset cancels within a molecule. For each test identity with at least two configurations, relative-energy MAE averages the absolute errors of all unordered energy differences, then averages identities equally. Selection success is the fraction of identities whose lowest hybrid-energy configuration lies within $1$~kcal/mol ($43.3641$~meV) of the identity's reference minimum. All $600$ COMP6 and $2{,}780$ QCML test identities qualify; MACE-OFF23 eligibility varies across reallocations and excludes singleton groups.

\begin{figure}[p]
\centering\includegraphics[width=\linewidth]{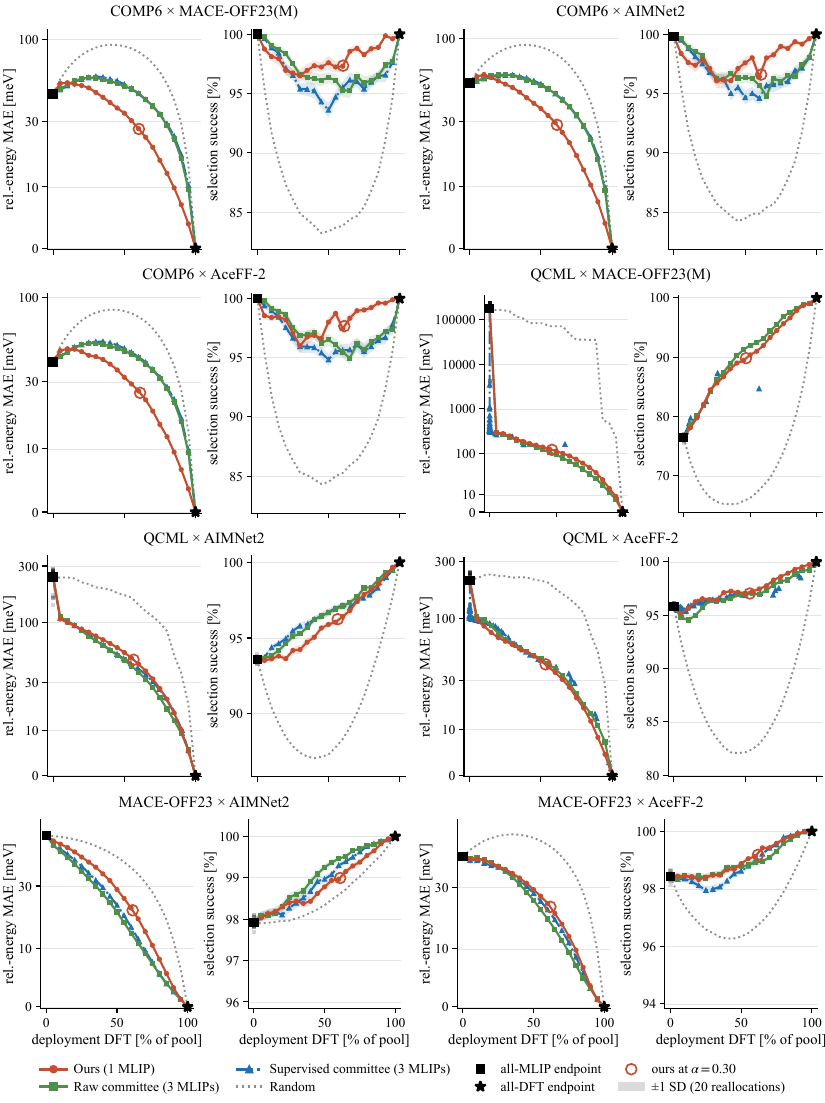}
\caption{\textbf{On COMP6, ours reaches a given relative-energy error and selection success with less deployment DFT than either committee; elsewhere the curves largely overlap.} Per cell, relative-energy MAE (asinh axis, linear below $10$ meV) and selection success against deployment DFT, with all-MLIP (square) and all-DFT (star) endpoints and a ring at ours, $\alpha=0.30$. Means over $20$ reallocations at $5\%$ DFT steps, each split at its nearest attainable point; bands $\pm1$ SD. The saved sweep has no oracle. Success axes are zoomed per cell.}
\label{fig:app-J2}
\end{figure}

\begin{figure}[htbp]
\centering\includegraphics[width=\linewidth]{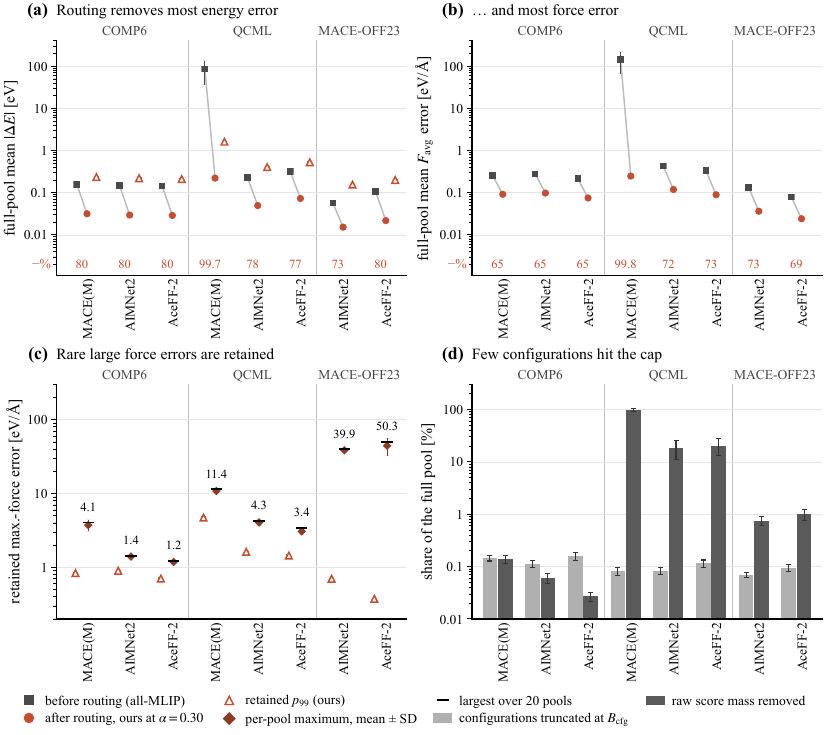}
\caption{\textbf{Routing at $\alpha=0.30$ removes most physical error, but rare large force errors remain.} (a) Full-pool mean aligned energy error and (b) mean atom-force error before routing and after routing with ours; numbers give the relative reduction (\%), triangles the retained 99th percentile. (c) Retained maximum-force error: 99th percentile, per-pool maximum (mean $\pm$ SD), and largest over $20$ pools. (d) Share of the pool truncated at $B_{\rm cfg}$, as configurations and as raw score mass. Log axes. Uncapped tails are outside the guarantee.}
\label{fig:tails}
\end{figure}

\paragraph{Findings.} At matched deployment DFT (Fig.~\ref{fig:app-J1}; procedure in Appendix~\ref{app:matched-resource}), ours lowers relative-energy MAE by $\MatchedRawMAEReduction$ and $\MatchedSupMAEReduction$~meV against the raw and supervised committees, better in $\MatchedMAEWinsRaw$ and $\MatchedMAEWinsSup$ of eight cells; the supervised mean is dominated by QCML $\times$ MACE-OFF23(M) ($-36.9$~meV). Selection success rises by $\MatchedRawSuccess$ and $\MatchedSupSuccess$ pp, better in $\MatchedSuccessWinsRaw$ and $\MatchedSuccessWinsSup$ of eight. The COMP6 curves separate clearly, while QCML and MACE-OFF23 largely overlap (Fig.~\ref{fig:app-J2}). At $\alpha=0.30$, ours reduces relative-energy MAE from the all-MLIP endpoint by at least $\MAEReductionMin\%$ in every cell, yet selection success falls below all-MLIP on the three COMP6 cells. Routing removes $73$--$99.7\%$ of the full-pool mean energy error and $65$--$99.8\%$ of the mean force error, but rare large force errors remain among retained configurations, up to $\RetainedForceMax$~eV/\AA{} on MACE-OFF23 $\times$ AceFF-2 (Fig.~\ref{fig:tails}). Truncation at $B_{\rm cfg}$ affects at most $\TruncationRateMax\%$ of configurations but removes $\QCMLMassRemoved\%$ of the raw score mass on QCML $\times$ MACE-OFF23(M); CRC controls only the capped loss.

\begin{figure}[htbp]
\centering\includegraphics[width=\linewidth]{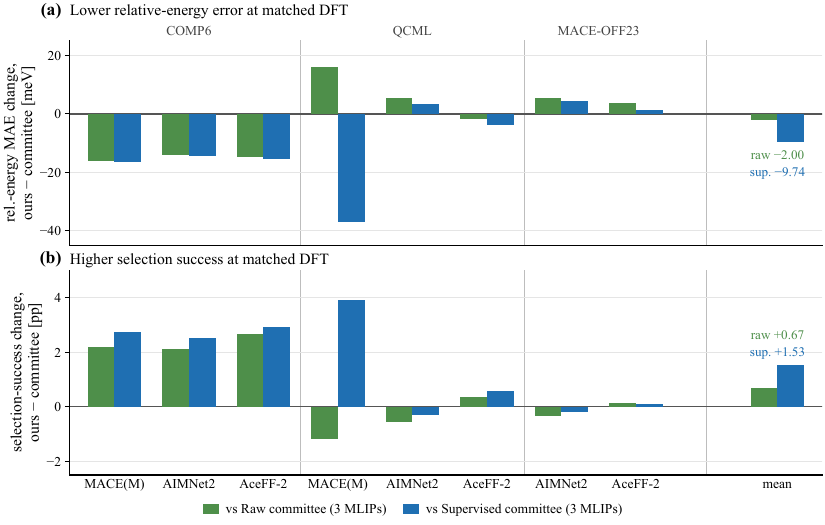}
\caption{\textbf{At matched deployment DFT, ours lowers relative-energy error and raises selection success on average, but not in every cell.} Per-cell difference, ours minus committee, against the raw (green) and supervised (blue) committees: (a) relative-energy MAE (lower is better); (b) selection success (higher is better). The right group is the equal-cell mean. Each value averages the difference of the $20$-reallocation mean curves over the cell's common attained deployment-DFT interval. Descriptive, not a new guarantee.}
\label{fig:app-J1}
\end{figure}

\subsection{Signed-residual cancellation}\label{app:cancellation}
\begin{figure}[htbp]
\centering\includegraphics[width=\linewidth]{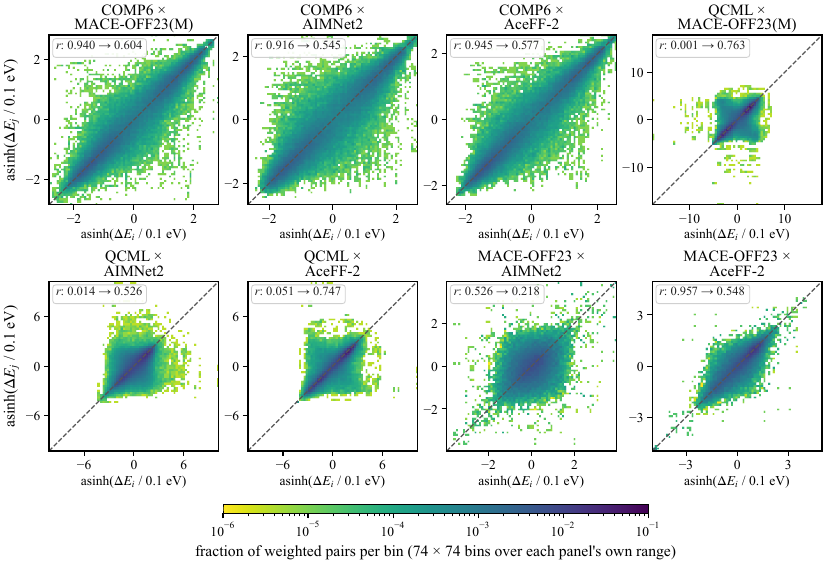}
\caption{\textbf{Signed residuals of one molecule's configurations are strongly correlated on COMP6 and MACE-OFF23 $\times$ AceFF-2, so energy differences benefit from cancellation that partial routing removes.} Weighted density of signed aligned residual pairs $(\Delta E_i,\Delta E_j)$ within an identity (up to $64$ pairs per eligible identity, both orientations), in $\operatorname{asinh}(\Delta E/0.1\,\mathrm{eV})$ coordinates; the dashed diagonal is exact cancellation. Annotations give the paired Pearson $r$ before routing $\to$ after ours at $\alpha=0.30$, averaged over $20$ reallocations. One shared log color scale of pair fraction per bin.}
\label{fig:cancellation}
\end{figure}

\begin{figure}[htbp]
\centering\includegraphics[width=\linewidth]{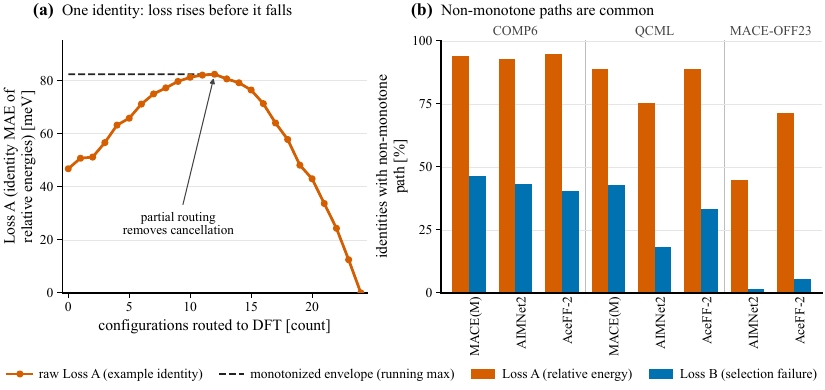}
\caption{\textbf{Partial routing can raise a molecule's relative-energy error before full DFT drives it to zero, and such paths are common.} (a) One saved COMP6 $\times$ AceFF-2 identity: Loss A (identity MAE of relative energies) and its running-maximum envelope against the number of configurations routed; an illustration, not a prevalence estimate. (b) Per cell, the percentage of identities whose routing path is non-monotone for Loss A and for Loss B (selection failure).}
\label{fig:app-J5}
\end{figure}

The MLIP residuals of one molecule's configurations often share sign, and same-sign residuals cancel in an energy difference. Routing one configuration to DFT sets its residual to zero and removes that cancellation. Before routing, the identity-balanced correlation of paired residuals is $0.92$--$0.96$ on COMP6 and MACE-OFF23 $\times$ AceFF-2, and routing with ours at $\alpha=0.30$ lowers it to $0.55$--$0.60$ (Fig.~\ref{fig:cancellation}). On QCML the all-MLIP correlation is near zero and rises after routing because the catastrophic residuals are removed. As a result, a molecule's relative-energy loss need not fall monotonically as DFT is added: $45$--$95\%$ of eligible identities have a non-monotone routing path for relative-energy MAE and $1$--$46\%$ for selection failure (Fig.~\ref{fig:app-J5}). These aggregate diagnostics are consistent with lost cancellation but do not identify the cause of each selection failure.

\FloatBarrier

\section{Certifying other losses directly}\label{app:other-losses}

Remark~\ref{rem:otherloss} allows any bounded monotone loss in place of $\ell$, so downstream objectives can be certified directly instead of through the proxy. We consider two identity-level losses on identities with at least two configurations. \emph{Loss A} is the relative-energy MAE of the identity, capped at the largest fitting-identity path loss, with budget $\epsilon_A$ a multiple of each cell's fitting all-MLIP mean. \emph{Loss B} is the indicator that the selected lowest-energy configuration lies more than $1$~kcal/mol above the true minimum, with a probability budget $\epsilon_B$. Because both can be non-monotone in the cutoff (Appendix~\ref{app:cancellation}), we compare \emph{M}, CRC on the running-maximum monotone envelope of the loss; \emph{W-max} and \emph{W-mean}, CRC with whole-identity routing by the maximum or mean configuration score; \emph{L}, Learn-then-Test on the raw loss over a fixed fitting-set grid with $\delta=0.10$; \emph{P}, the original proxy rule at $\alpha=0.30$, which carries no downstream certificate; and \emph{N}, a naive empirical cutoff with no certificate. M and W bound the expectation, whereas L holds with probability at least $0.90$ over the calibration draw, so their costs are not directly comparable. DFT percentages here count eligible non-singleton configurations.

\begin{figure}[htbp]
\centering\includegraphics[width=\linewidth]{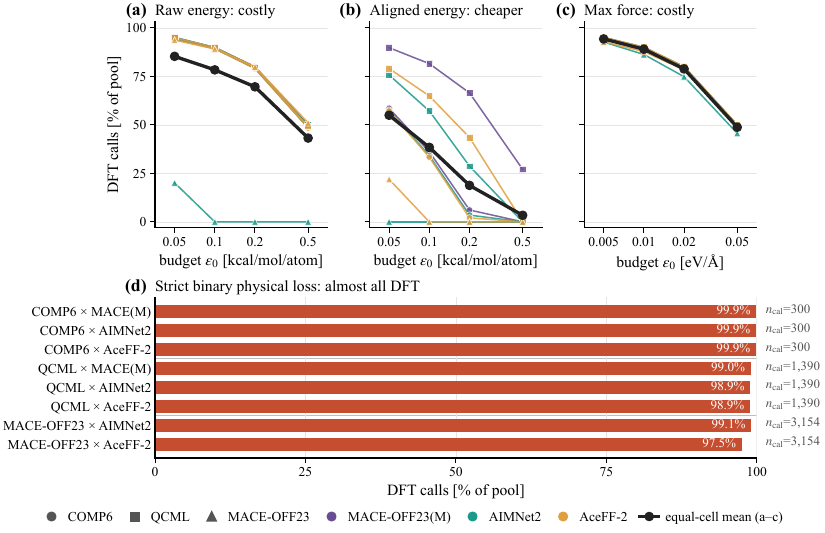}
\caption{\textbf{Strict physical-unit budgets cost almost all of the DFT savings.} (a--c) DFT calls against an absolute budget for raw and aligned energy (kcal/mol/atom) and maximum force error (eV/\AA), with the same structural score and CRC rule; thin lines are cells, the bold line the equal-cell mean of $20$-reallocation means. (d) A strict binary loss (aligned $|\Delta E|>1$ kcal/mol or maximum force $>0.10$ eV/\AA; budget $0.10$ violations per identity) at each cell's calibration size. Thresholds are examples, not universal tolerances.}
\label{fig:app-K1}
\end{figure}

\begin{figure}[htbp]
\centering\includegraphics[width=\linewidth]{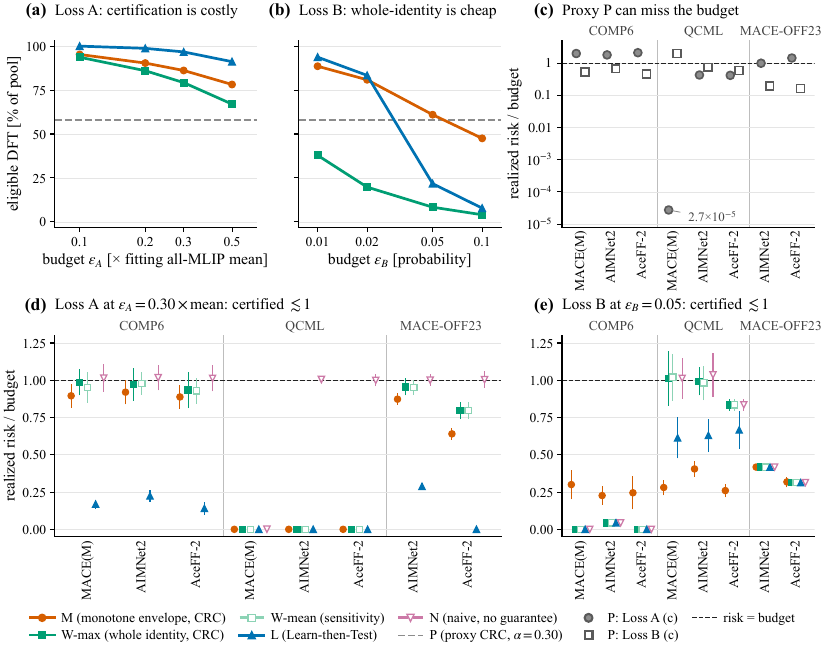}
\caption{\textbf{The proxy guarantee does not carry over to downstream losses, and direct control has its own cost.} (a, b) Eligible DFT against the budget for Loss A and Loss B under M, W-max, and L; the dashed line is the uncertified proxy rule P at $\alpha=0.30$. Equal-cell means of $20$-repeat means. (c) Realized risk/budget of P per cell (log axis); the tiny QCML $\times$ MACE-OFF23(M) Loss-A ratio reflects its extreme fitting anchor. (d, e) Per-cell realized risk/budget of every method at the main budgets, mean $\pm$ SD over $20$ repeats. L is high-probability; M and W are expectation-level.}
\label{fig:app-K2}
\end{figure}

\begin{figure}[p]
\centering\includegraphics[width=\linewidth]{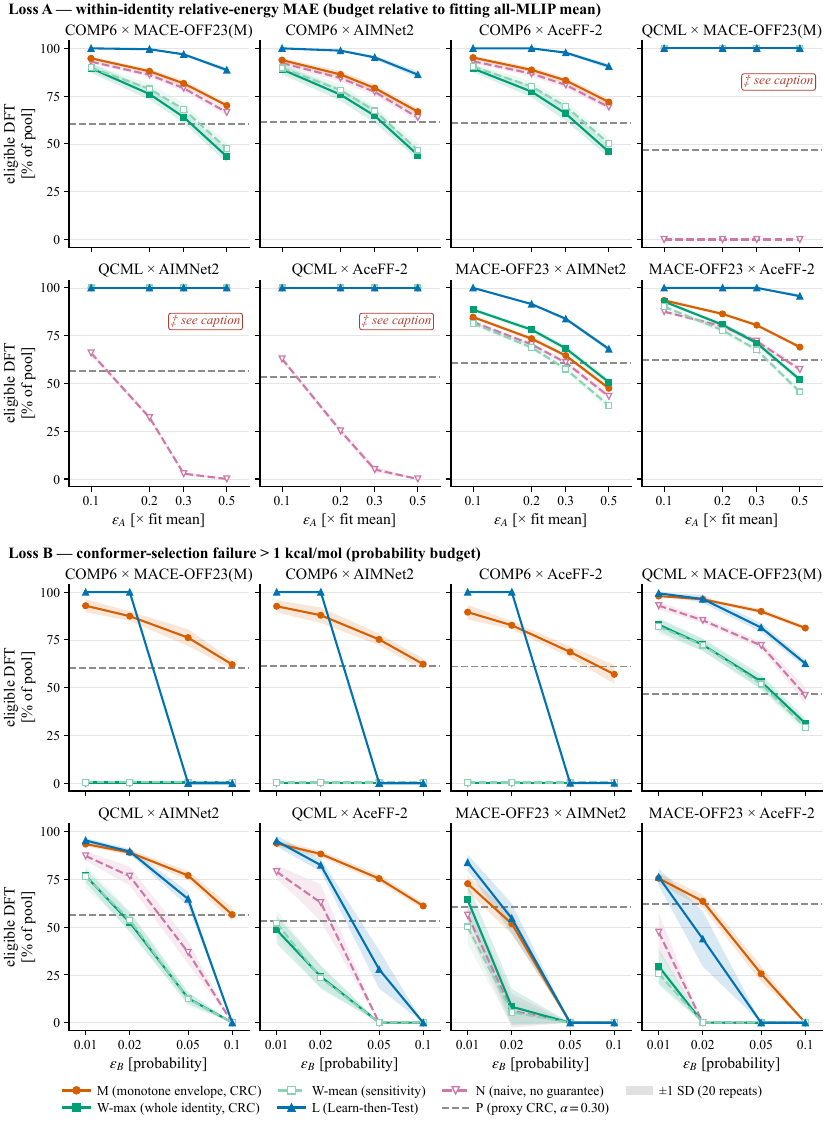}
\caption{\textbf{Downstream certification cost depends strongly on the cell and the objective.} Eligible DFT against the budget in every cell, Loss A (top two rows) and Loss B (bottom two), for M, W-max, W-mean, L, and N (mean $\pm1$ SD over $20$ repeats); P is a horizontal reference. \ddag~Data issue, plotted without alteration: the saved QCML $\times$ MACE-OFF23(M) Loss-A fitting mean is $14{,}761$~eV (main budget $4{,}428.5$~eV), and on all three QCML Loss-A panels every certified method routes $100\%$. Budgets are relative to fitting anchors, not a common absolute accuracy.}
\label{fig:app-K3}
\end{figure}

\paragraph{Findings.} Strict physical-unit losses, certified with the same score and CRC rule, cost almost all savings (Fig.~\ref{fig:app-K1}): at the smallest budgets the equal-cell mean DFT is $\PhysAlignedDFT\%$ for aligned energy ($0.05$~kcal/mol/atom), $85.3\%$ for raw energy, and $94.3\%$ for maximum force ($0.005$~eV/\AA), and a strict binary loss needs $97.5$--$99.9\%$. The proxy guarantee does not transfer downstream: P exceeds the Loss-A budget ($0.3\times$ the fitting mean) in five of eight cells, with ratios $1.01$--$2.14$, and the Loss-B budget ($0.05$) in one (Fig.~\ref{fig:app-K2}c). Direct certification costs depend on the objective (Fig.~\ref{fig:app-K2}a,\,b): at the main Loss-A budget, M, W-max, and L route $86.1\%$, $79.3\%$, and $96.7\%$, while at $\epsilon_B=0.05$ they route $61.0\%$, $8.4\%$, and $21.8\%$; W-max is cheap partly because all-MLIP already meets the selection budget in several cells. Expectation-level methods can exceed the budget in individual repeats (Fig.~\ref{fig:app-K2}d,\,e). Relative Loss-A budgets correspond to different absolute accuracies across cells. In particular, the saved QCML $\times$ MACE-OFF23(M) fitting mean is $14{,}761$~eV, so its main budget is $4{,}428.5$~eV, and on all three QCML Loss-A panels every certified method routes $100\%$ (Fig.~\ref{fig:app-K3}). We report these values as saved; the export does not establish the cause of the extreme scale.

\FloatBarrier

\section{Label cost and transfer}\label{app:economics}

\begin{figure}[htbp]
\centering\includegraphics[width=\linewidth]{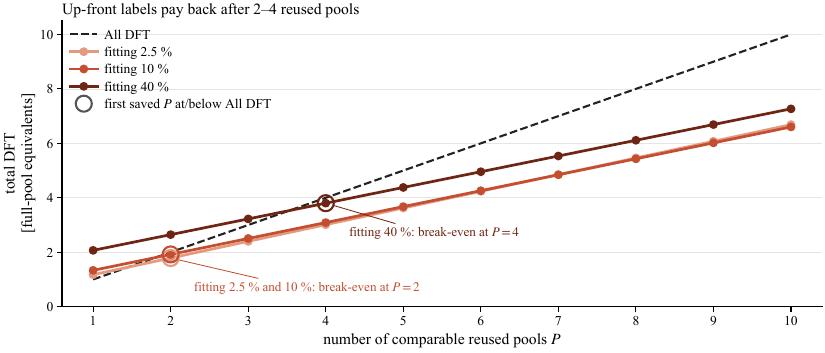}
\caption{\textbf{Up-front labels pay back after 2--4 reused pools.} Total DFT in full-pool equivalents (fitting and calibration labels up front, plus routed deployment labels) against the number $P$ of comparable reused pools, for fitting at $2.5\%$, $10\%$, and $40\%$ of identities and for all-DFT ($y=P$). Equal-cell means, each cell normalized by one deployment pool; rings mark the first saved integer $P$ at or below all-DFT. Cost-model projections in configuration counts, not measured wall time.}
\label{fig:app-L1}
\end{figure}

\begin{figure}[htbp]
\centering\includegraphics[width=\linewidth]{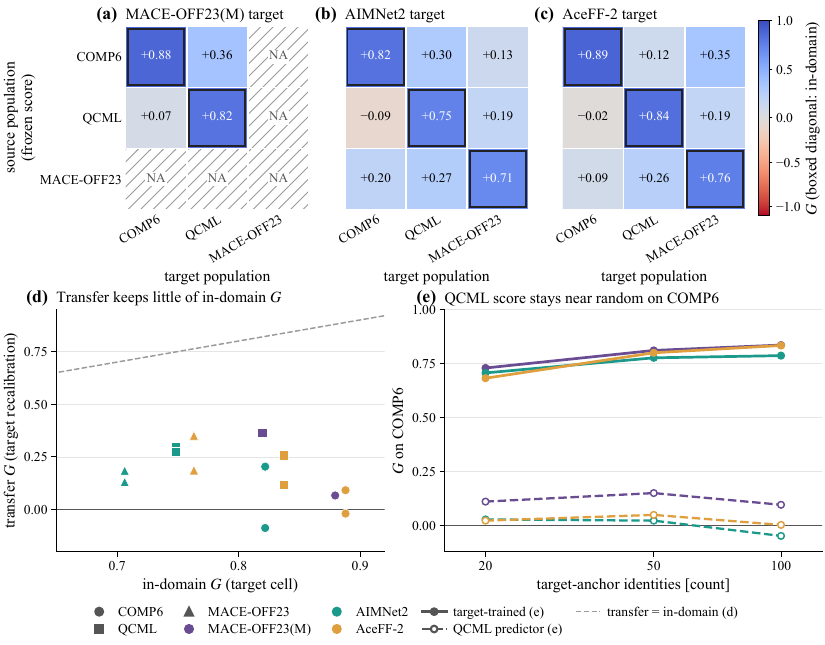}
\caption{\textbf{A frozen structural score transfers poorly across populations.} (a--c) $G$ of the frozen score from each source population (rows) on each target population (columns), per target MLIP, after target recalibration; boxed diagonals are in-domain, NA marks non-cells. (d) Transfer $G$ against the target's in-domain $G$ for all $14$ directed transfers (dashed: equality). (e) Anchored QCML $\to$ COMP6: $G$ against the number of target-anchor identities for the target-trained (solid) and frozen QCML (dashed) predictors. Target anchors and calibration labels are charged, so this is not zero-label deployment.}
\label{fig:app-L2}
\end{figure}

The reported savings are per reused pool and exclude the labels of $\Dfit$ and $\Dcal$. Charging them up front, and counting each cell in full-pool equivalents, the fitted selector breaks even after $P=2$ comparable reused pools with $2.5\%$ or $10\%$ of identities for fitting and after $P=4$ with $40\%$ (Fig.~\ref{fig:app-L1}); for a single pool it costs $1.17$--$2.07$ full-pool equivalents, more than labeling the pool outright. Reuse assumes comparable pools, and the predictor must be fitted per population: a frozen structural score moved to another population keeps mean $G=\TransferMeanG$ (range $\TransferMinG$ to $\TransferMaxG$, positive in $\TransferPositive$ of $14$ directed transfers) after target recalibration, against $\TransferInDomainG$ in domain, and the QCML predictor anchored on COMP6 stays at $G\le0.15$ (Fig.~\ref{fig:app-L2}).

\FloatBarrier

\section{Limitations}\label{app:limitations}

\textbf{Scope of the guarantee.} As is standard for conformal risk control, Proposition~\ref{thm:crc} is a marginal, expectation-level bound that holds within the dataset--MLIP cell used for fitting; a new population or target MLIP requires refitting and recalibration (Appendix~\ref{app:economics}). The bound applies to the capped loss, and $B_{\mathrm{cfg}}$ can be set to an application tolerance. Where rare chemical families or single calibration draws must be protected, high-probability or group-conditional variants apply at additional DFT cost (Appendix~\ref{app:other-losses}).

\textbf{Cost accounting.} Reported savings are per deployment pool. The one-time fitting and calibration labels are amortized after two to four comparable pools (Appendix~\ref{app:economics}), so the method targets repeated screening of a population.

\textbf{Reference mismatch.} The residuals include differences between electronic-structure levels, which is the setting the method is designed for. When references match, committee disagreement is competitive, and the hybrid score combines both signals (Appendix~\ref{app:mechanism}).

\textbf{Downstream objectives.} The proxy loss does not certify relative energies or conformer selection. Because the framework accepts any bounded monotone loss, these objectives can be certified directly (Appendix~\ref{app:other-losses}).

\textbf{Domain.} The descriptor covers isolated molecules built from ten elements; periodic systems require extending its distance definitions.

\section{Broader Impacts}\label{app:broader-impacts}

Routing lowers the DFT cost of repeated molecular screening. It also replaces empirically chosen uncertainty thresholds with a stated, auditable loss and budget, which may help pretrained MLIPs enter workflows that require documented error control. Two risks follow. First, cheaper screening can enlarge campaigns, so total compute need not fall. Second, if routed labels are used to retrain the MLIP, regions the selector ranks low may remain unlabeled, and retraining requires refitting the predictor and recalibrating the threshold. The largest practical risk is overreading the guarantee as assurance of stable trajectories, correct conformer energies, or the absence of rare large errors, none of which it covers. Reports of routed results should state the controlled loss, its caps, and the calibration population.

%% file: tables/T_main.tex
\setlength{\tabcolsep}{3pt}
\renewcommand{\arraystretch}{1.12}
\begin{tabular}{lcrrr rrr}
\toprule
& & & \multicolumn{2}{c}{at $\alpha=0.30$} & \multicolumn{3}{c}{committee vs.\ ours at matched resources} \\
\cmidrule(lr){4-5}\cmidrule(lr){6-8}
Method & \shortstack{MLIP\\evals} & $G$ & DFT [\%] & $\rho$ & \shortstack{DFT saved\\{[pp of pool]}} & \shortstack{rel.-energy MAE\\{[meV]}} & \shortstack{selection success\\{[pp]}} \\
\midrule
Ours                  & 1 & \textbf{0.808} & \textbf{57.8} & 0.289 & --- & --- & --- \\
Raw committee         & 3 & 0.497 & 61.9 & 0.282 & $+3.0$ & $-2.0$ & $+0.7$ \\
Supervised committee  & 3 & 0.341 & 67.5 & 0.251 & $+4.0$ & $-9.7$ & $+1.5$ \\
Random                & 1 & 0.000 & 71.6 & 0.284 & --- & --- & --- \\
\midrule
Oracle (retrospective) & --- & 1.000 & 54.4 & 0.288 & --- & --- & --- \\
\bottomrule
\end{tabular}

%% file: tables/T5_score_summary.tex
\setlength{\tabcolsep}{3pt}
\renewcommand{\arraystretch}{1.12}
\begin{tabular}{@{}lrrr rrr@{}}
\toprule
& \multicolumn{3}{c}{primary scores} & \multicolumn{3}{c}{label-free controls} \\
\cmidrule(lr){2-4}\cmidrule(lr){5-7}
Cell & Ours & Raw & Supervised & kNN-1 & kNN-5 & Max-force \\
\midrule
COMP6 $\times$ MACE-OFF23(M) & \textbf{0.879} & 0.223 & \underline{0.274} & 0.261 & 0.267 & 0.134 \\
COMP6 $\times$ AIMNet2 & \textbf{0.822} & 0.237 & \underline{0.255} & 0.236 & 0.244 & 0.131 \\
COMP6 $\times$ AceFF-2 & \textbf{0.888} & 0.136 & \underline{0.183} & 0.226 & 0.232 & 0.079 \\
QCML $\times$ MACE-OFF23(M) & \underline{0.820} & \textbf{0.910} & -0.278 & 0.362 & 0.385 & 0.245 \\
QCML $\times$ AIMNet2 & \textbf{0.748} & \underline{0.529} & 0.395 & 0.371 & 0.389 & 0.322 \\
QCML $\times$ AceFF-2 & \textbf{0.837} & \underline{0.523} & 0.317 & 0.333 & 0.361 & 0.163 \\
MACE-OFF23 $\times$ AIMNet2 & 0.706 & \underline{0.716} & \textbf{0.763} & 0.471 & 0.471 & 0.419 \\
MACE-OFF23 $\times$ AceFF-2 & \underline{0.763} & 0.705 & \textbf{0.815} & 0.246 & 0.272 & 0.315 \\
\midrule
Mean & \textbf{0.808} & \underline{0.497} & 0.341 & 0.313 & 0.327 & 0.226 \\
\bottomrule
\end{tabular}

%% file: tables/T_matched_resources.tex
\setlength{\tabcolsep}{3pt}
\renewcommand{\arraystretch}{1.18}
\begin{tabular*}{\linewidth}{@{\extracolsep{\fill}}llrrrrrr@{}}
\toprule
\multicolumn{8}{@{}l}{\textbf{(a) Matched-resource comparisons}} \\[3pt]
Dataset & MLIP & \multicolumn{2}{c}{\shortstack{DFT saving $\uparrow$\\{[pp]}}} & \multicolumn{2}{c}{\shortstack{MAE change $\downarrow$\\{[meV]}}} & \multicolumn{2}{c}{\shortstack{Success change $\uparrow$\\{[pp]}}} \\
\cmidrule(lr){3-4}\cmidrule(lr){5-6}\cmidrule(l){7-8}
 & & Raw & Sup. & Raw & Sup. & Raw & Sup. \\
\midrule
COMP6 & MACE-OFF23(M) & +5.85 & +5.28 & -16.12 & -16.60 & +2.19 & +2.73 \\
 & AIMNet2 & +5.01 & +4.77 & -14.13 & -14.40 & +2.10 & +2.51 \\
 & AceFF-2 & +6.47 & +5.94 & -14.63 & -15.43 & +2.67 & +2.91 \\
\addlinespace[4pt]
QCML & MACE-OFF23(M) & -2.51 & +8.45 & +16.13 & -36.87 & -1.19 & +3.93 \\
 & AIMNet2 & +3.50 & +4.24 & +5.56 & +3.25 & -0.55 & -0.32 \\
 & AceFF-2 & +5.49 & +5.69 & -1.80 & -3.69 & +0.34 & +0.58 \\
\addlinespace[4pt]
MACE-OFF23 & AIMNet2 & -0.04 & -0.94 & +5.40 & +4.30 & -0.33 & -0.21 \\
 & AceFF-2 & +0.39 & -1.34 & +3.57 & +1.50 & +0.15 & +0.09 \\
\midrule
\textbf{Mean} & & \textbf{+3.02} & \textbf{+4.01} & \textbf{-2.00} & \textbf{-9.74} & \textbf{+0.67} & \textbf{+1.53} \\
\bottomrule
\end{tabular*}
\par\vspace{8pt}
\begin{tabular*}{\linewidth}{@{\extracolsep{\fill}}llcc@{}}
\toprule
\multicolumn{4}{@{}l}{\textbf{(b) Common attained support}} \\[3pt]
Dataset & MLIP & Realized loss $I_\rho$ & Deployment DFT $I_{\rm DFT}$ [\%] \\
\midrule
COMP6 & MACE-OFF23(M) & [0.048, 0.479] & [48.17, 92.28] \\
 & AIMNet2 & [0.047, 0.480] & [48.22, 92.78] \\
 & AceFF-2 & [0.047, 0.478] & [49.95, 92.63] \\
\addlinespace[4pt]
QCML & MACE-OFF23(M) & [0.038, 0.476] & [27.06, 87.11] \\
 & AIMNet2 & [0.042, 0.492] & [38.79, 92.22] \\
 & AceFF-2 & [0.048, 0.487] & [38.39, 90.11] \\
\addlinespace[4pt]
MACE-OFF23 & AIMNet2 & [0.027, 0.459] & [39.90, 94.57] \\
 & AceFF-2 & [0.021, 0.442] & [41.23, 94.94] \\
\bottomrule
\end{tabular*}

%% file: tables/T4_headroom.tex
\setlength{\tabcolsep}{3pt}
\renewcommand{\arraystretch}{1.12}
\begin{tabular}{llrrrrrrr}
\toprule
Dataset & MLIP & Gini & Top 10\% & $m_{\max}$ & $B_{\rm cfg}$ & $B$ & Med. $|\Delta E|$ & $p_{99}|\Delta E|$ \\
\midrule
COMP6 & MACE-OFF23(M) & 0.18 & 0.16 & 24 & 3.4 & 45.4 & 139 & 516 \\
 & AIMNet2 & 0.16 & 0.15 & 24 & 2.2 & 46.4 & 132 & 474 \\
 & AceFF-2 & 0.16 & 0.15 & 24 & 2.1 & 47.4 & 132 & 443 \\
\addlinespace[2pt]
QCML & MACE-OFF23(M) & 1.00 & 1.00 & 9 & 48.2 & 158.5 & 485 & 3{,}895 \\
 & AIMNet2 & 0.45 & 0.40 & 9 & 34.3 & 122.6 & 126 & 839 \\
 & AceFF-2 & 0.46 & 0.40 & 9 & 11.0 & 42.4 & 202 & 1{,}151 \\
\addlinespace[2pt]
MACE-OFF23 & AIMNet2 & 0.30 & 0.42 & 229 & 19.3 & 210.2 & 41 & 277 \\
 & AceFF-2 & 0.30 & 0.45 & 229 & 25.0 & 273.3 & 82 & 396 \\
\bottomrule
\end{tabular}